\documentclass[sigconf]{acmart}
\usepackage{subcaption}
\usepackage{multirow}
\usepackage{pifont}
\renewcommand\footnotetextcopyrightpermission[1]{} 
\AtBeginDocument{%
  }

\setcopyright{acmlicensed}
\copyrightyear{2018}
\acmYear{2018}
\acmDOI{XXXXXXX.XXXXXXX}
\begin{document}

\title{TimeCatcher: A Variational Framework for Volatility-Aware Forecasting of Non-Stationary Time Series}

\author{Zhiyu Chen}
\email{2023080136002@std.uestc.edu.cn}
\orcid{0009-0003-7775-7802}
\affiliation{%
  \institution{University of Electronic Science and Technology of China}
  \city{Chengdu}
  \state{Sichuan}
  \country{China}
}

\author{Minhao Liu}
\authornote{Corresponding author.}
\email{minhaoliu@uestc.edu.cn}
\affiliation{%
  \institution{Shenzhen Institute for Advanced Study, UESTC}
  \city{Shenzhen}
  \state{Guangdong}
  \country{China}
}

\author{Yanru Zhang}
\email{yanruzhang@uestc.edu.cn}
\affiliation{%
  \institution{University of Electronic Science and Technology of China}
  \city{Chengdu}
  \state{Sichuan}
  \country{China}
}
\affiliation{%
  \institution{Shenzhen Institute for Advanced Study, UESTC}
  \city{Shenzhen}
  \state{Guangdong}
  \country{China}
}

\renewcommand{\shortauthors}{Zhiyu Chen et al.}

\begin{abstract}
  Recent lightweight MLP-based models have achieved strong performance in time series forecasting by capturing stable trends and seasonal patterns. However, their effectiveness hinges on an implicit assumption of local stationarity assumption, making them prone to errors in long-term forecasting of \textbf{highly non-stationary series}, especially when abrupt fluctuations occur, a common challenge in domains like web traffic monitoring. To overcome this limitation, we propose \textbf{TimeCatcher}, a novel \textbf{Volatility-Aware Variational Forecasting framework}. TimeCatcher extends linear architectures with a variational encoder to capture latent dynamic patterns hidden in historical data and a volatility-aware enhancement mechanism to detect and amplify significant local variations. Experiments on nine real-world datasets from traffic, financial, energy, and weather domains show that TimeCatcher consistently outperforms state-of-the-art baselines, with particularly large improvements in long-term forecasting scenarios characterized by high volatility and sudden fluctuations. Our code is available at https://github.com/ColaPrinceCHEN/TimeCatcher.
\end{abstract}




\maketitle

\section{Introduction}

Time series forecasting plays a critical role in real-world applications such as web traffic monitoring \cite{10.1145/3696410.3714647,  10.1145/3701716.3715582}, financial trading \cite{10.1145/3701716.3715216, 10.1145/3701716.3717509, 10.1145/3701716.3715551}, energy load management \cite{wang2023probabilistic}, and transportation planning \cite{10.1145/3701716.3715202, 10.1145/3701716.3715464}. In these scenarios, accurate long-term prediction and timely response to abrupt fluctuations, such as sudden traffic spikes or market crashes, are both essential.

\begin{figure}[htbp]
    \centering
    \begin{subfigure}[b]{0.22\textwidth}
        \includegraphics[width=\linewidth]{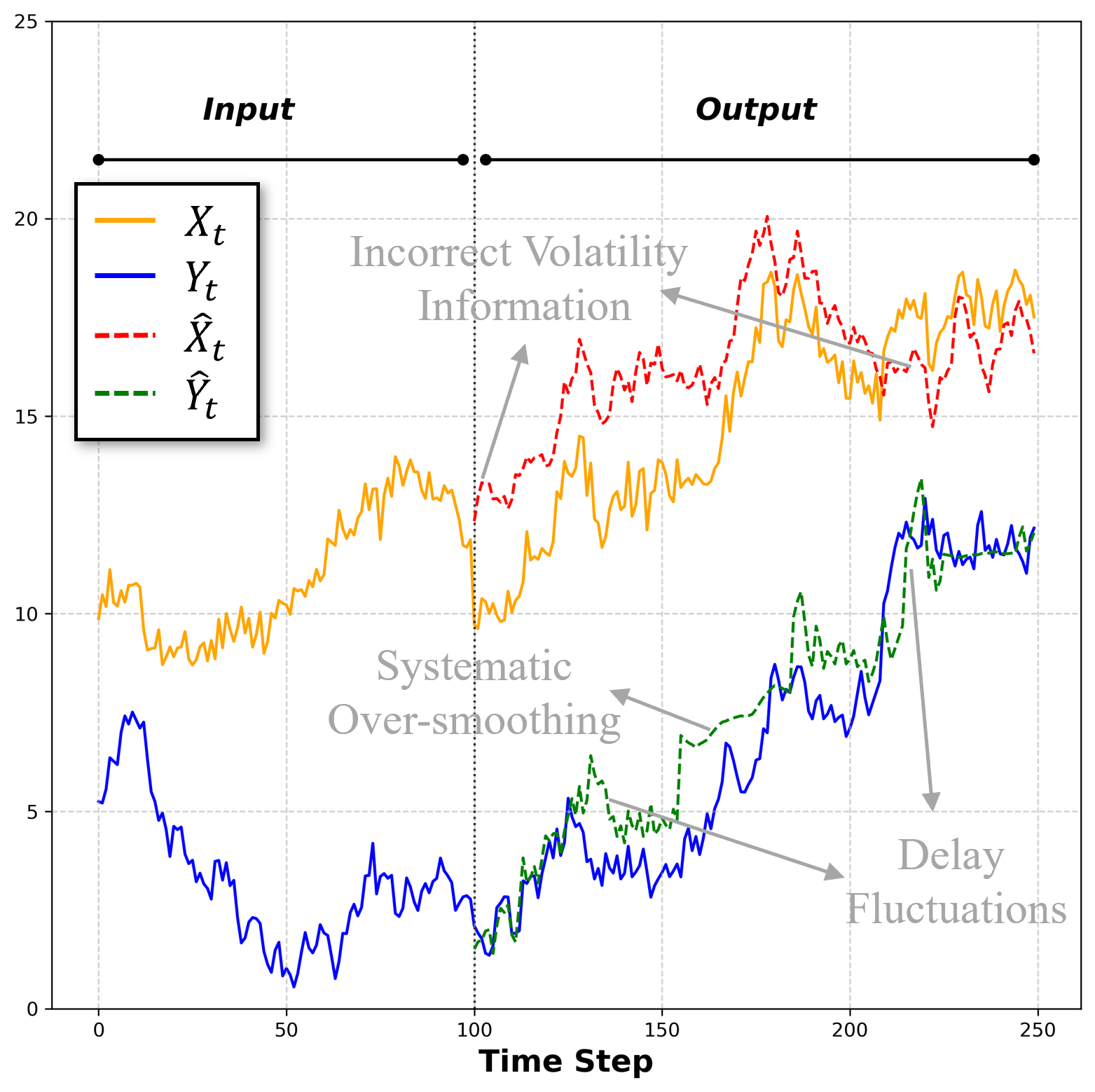}
        \caption{Traditional Model}
        \label{fig:sub1}
    \end{subfigure}
    \hfill
    \begin{subfigure}[b]{0.22\textwidth}
        \includegraphics[width=\linewidth]{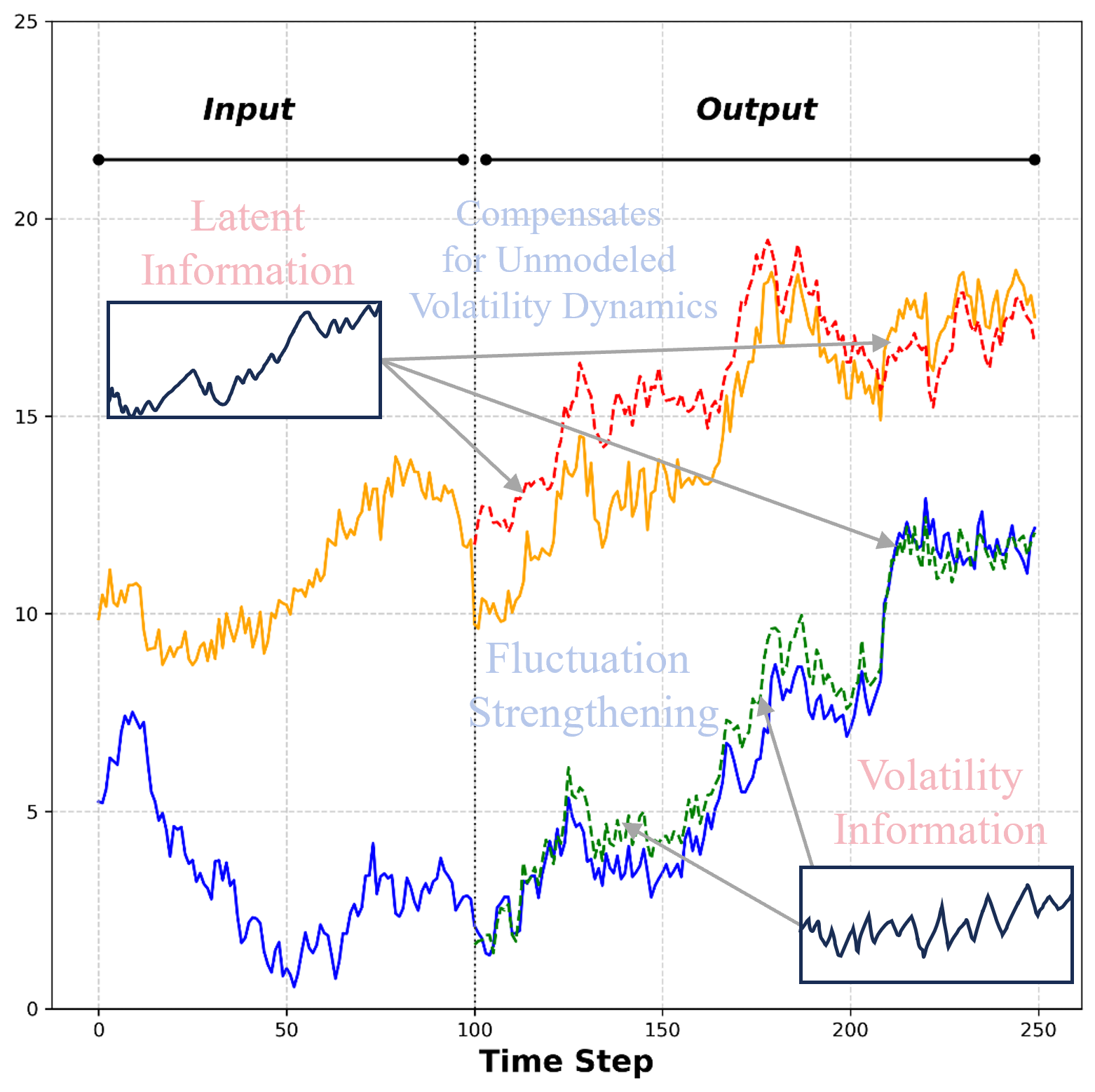}
        \caption{TimeCatcher}
        \label{fig:sub2}
    \end{subfigure}
    \caption{Visualize the performance of predicting non-stationary sequences, where $X_t$ and $Y_t$ are input sequences and $\hat{X}_t$ and $\hat{Y}_t$ are predicted sequences. By capturing latent and Volatility information, TimeCatcher can significantly improve forecasting performance.}
    \label{fig:both}
\end{figure}

Recent Transformer-based models \cite{woo2022etsformer, liu2022pyraformer, zhou2022fedformer} can model long-range dependencies and complex dynamics. However, they suffer from high parameter counts and slow inference, making them impractical for resource-constrained or real-time deployment settings. In contrast, lightweight models based on multilayer perceptrons (MLPs) \cite{zeng2023transformers, zhang2024neural, oreshkin2019n} have demonstrated strong performance in capturing stable trends and seasonal patterns, thanks to their simple and computationally efficient architecture. Nevertheless, they often rely on implicit assumptions of local stationarity, leading to systematic over-smoothing of abrupt changes and insufficient responsiveness to critical events in highly non-stationary time series (\textbf{Figure 1}). This raises a natural question: \emph{Can MLP-based models be adapted to capture volatility and abrupt changes in highly non-stationary series without compromising their efficiency?}

To address this challenge, we propose \textbf{TimeCatcher}, a novel Volatility-Aware Variational Forecasting Framework. TimeCatcher builds upon an MLP-based forecasting backbone that explicitly captures, amplifies, and predicts volatility in time series without sacrificing the computational efficiency typical of linear models. Our core idea is to decompose the forecasting distribution into three additive residual components: (1) a trend term derived from the raw input, preserving global trends and seasonal patterns;
(2) a latent dynamic term modeled by a Variational Autoencoder (VAE) encoder, which learns the underlying data distribution and encodes temporal dynamics into a compact latent space, enabling the model to inherit statistical regularities from training data;
(3) a volatility enhancement term, implemented via a lightweight volatility-aware enhancement module that adaptively identifies and amplifies significant local variations using a learnable weighting mechanism, thereby strengthening the representation of abrupt changes while maintaining overall stability. In summary, our main contributions are as follows:
\begin{itemize}
\item We propose \textbf{TimeCatcher}, a lightweight hybrid forecasting framework that integrates VAE-based latent modeling with a volatility-aware module, effectively balancing model efficiency and sensitivity to non-stationary dynamics.
\item We design a \textbf{volatility enhancement module} that detects and amplifies local variations, significantly improving the model's responsiveness to abrupt fluctuations in non-stationary time series.
\item We conduct extensive experiments and visualization studies demonstrating the compactness and structure of the learned latent representations. Results show that TimeCatcher outperforms state-of-the-art methods in long-term forecasting, particularly in high-volatility domains such as electricity load and financial markets.
\end{itemize}

\section{Related Work}
\subsection{Time Series Forecasting}
The field of time series forecasting has undergone a paradigm shift in recent years, transitioning from traditional statistical methods to deep learning. Traditional models like ARIMA \cite{Anderson1976TimeSeries2E} and STL \cite{cleveland1990terpenning} effectively analyze periodic and trend patterns in time series data but struggle with nonlinear dynamics. Deep learning models can be broadly categorized into four types: CNN-based, RNN-based, Transformer-based, and MLP-based. Typically, CNN-based models employ one-dimensional dilated convolutions to capture long-term dependencies in sequences, yet their performance is constrained by limited receptive fields \cite{wang2023micn, li2021modeling}. TCNs improve local pattern extraction but remain constrained in capturing long-term dependencies \cite{hewage2020temporal, franceschi2019unsupervised, bai2018empirical}. RNN-based methods inherently process sequential data through their recurrent structure, yet issues like vanishing or exploding gradients make learning very long-term dependencies practically challenging \cite{wen2017multi, du2020multivariate, gangopadhyay2021spatiotemporal, redhu2023short}.

Recently, following their revolutionary success in natural language processing, Transformer models and their self-attention mechanisms have been introduced to time series forecasting \cite{lim2021temporal, du2023preformer}. Models like Informer \cite{zhou2021informer}, Autoformer \cite{wu2021autoformer}, and iTransformer directly model global dependencies between any two points in a sequence via attention mechanisms, demonstrating strong performance across multiple long-term forecasting tasks \cite{liu2023itransformer}. PatchTST's batch processing technology also offers an innovative approach to time series analysis \cite{nie2022time}. However, transformer-based methods exhibit high computational complexity and are prone to overfitting in time series forecasting tasks.

Against this backdrop, simple MLP-based architectures have regained attention. Studies like N-BEATS and DLinear demonstrate that carefully designed MLPs can match or even surpass the performance of many complex models \cite{oreshkin2019n, zeng2023transformers}. The core of such models lies in directly applying linear layers or MLPs to operate on the sequence dimension or feature dimension of time series data, offering high computational efficiency and interpretability. Our work continues this pursuit of efficient, concise architectures, but differs by focusing on addressing the specific shortcomings of MLPs and existing models in fluctuation prediction.

\begin{figure*}[t]
  \includegraphics[width=\textwidth]{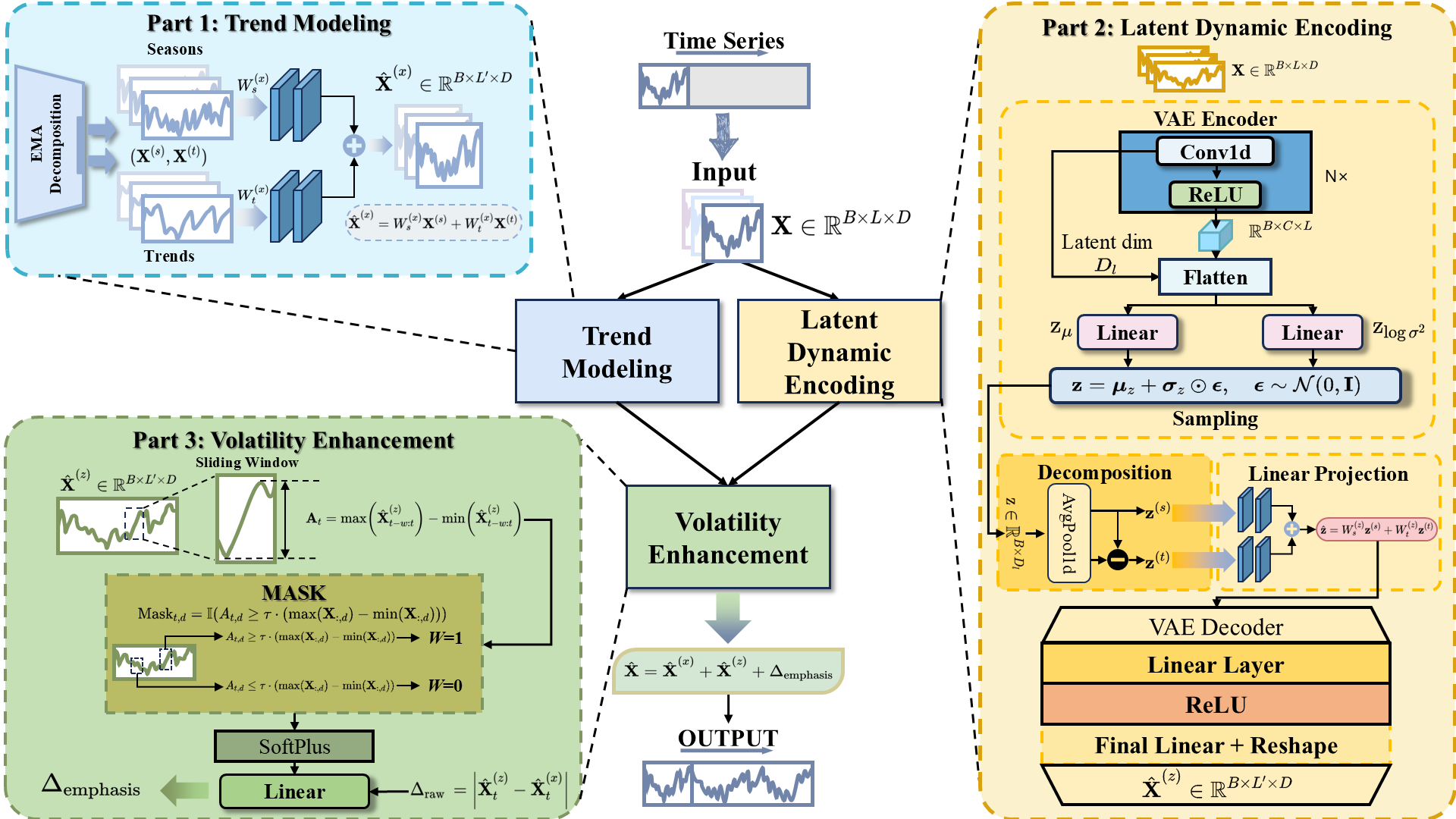}
  \caption{The TimeCatcher framework. The input time series is processed through three additive residual pathways: (1) \textbf{Part 1: Trend Modeling} processes the raw input to preserve global patterns; (2) \textbf{Part 2: Latent Dynamic Encoding} uses a VAE-based encoder to capture underlying temporal structures in a compact latent space; (3) \textbf{Part 3: Volatility Enhancement} adaptively amplifies local fluctuations via a lightweight module with learnable weights. The final prediction is formed by summing the outputs of all three pathways. This decomposition enables TimeCatcher to simultaneously achieve high efficiency, inherit historical dynamics, and respond sensitively to abrupt changes in non-stationary series.}
  \label{fig:Structure}
\end{figure*}

\subsection{Variational Autoencoder}
Variational autoencoders (VAEs) serve as powerful generative models primarily employed in time series analysis for learning smooth latent representations and performing probabilistic predictions \cite{girin2020dynamical}. Their application in time series typically follows this paradigm: an encoder maps input sequences to a probability distribution over the latent space, from which latent variables are sampled. Finally, a decoder reconstructs the input or predicts future values based on these latent variables. Temporal VAEs learn dynamic latent representations of sequences by employing RNNs or TCNs as encoders and decoders \cite{li2024cumulative, dissanayaka2024temporal}. These studies demonstrate that the latent space of VAEs effectively captures abstract sequence features such as operational patterns and trends while filtering out some noise, providing more robust representations for prediction. However, while VAEs excel at learning smooth latent representations that conform to a prior distribution, this very property may lead to over-smoothing. Addressing both the over-smoothing issue in VAEs and the aforementioned limitations of complex models in capturing sudden fluctuations, this paper proposes a novel collaborative architecture. It enhances the VAE encoder by integrating a lightweight “expert” module specifically designed to compensate for its deficiencies in capturing abrupt information.

\section{Methodology}
Time series forecasting aims to predict future observations $\hat{\mathbf{X}} \in \mathbb{R}^{B \times L^{\prime} \times D}$ given a historical sequence $\mathbf{X} \in \mathbb{R}^{B \times L \times D}$, where $B$ denotes the batch size, $L$ and $L^{\prime}$ are the lengths of input and prediction horizons, and $D$ is the feature dimension. Traditional forecasting models primarily focus on modeling surface-level temporal dependencies, often struggling to capture latent dynamics and abrupt structural changes that are crucial in real-world scenarios. To address these challenges, we design TimeCatcher, a three-stage forecasting framework that integrates: (1) a VAE-based latent state modeling module to learn hidden temporal representations, (2) a volatility emphasis module to enhance the model’s sensitivity to sudden fluctuations, and (3) a source-space residual forecasting module that preserves deterministic trends. Together, these components enable TimeCatcher to jointly capture deterministic, latent, and volatility-aware dynamics for robust and accurate long-term forecasting.

\subsection{Volatility-Aware Variational Forecasting Framework}
Forecasting abrupt, non-stationary time series is fundamentally challenging because the predictive signal exists at multiple semantic levels: low-frequency deterministic trends, latent dynamic dependencies, and transient structural shifts. Most existing approaches, however, treat these components in a monolithic fashion — leading to over-smoothing, insufficient interpretability, and degraded robustness during regime changes.

To overcome these limitations, TimeCatcher is designed as a \textbf{volatility-aware variational forecasting framework} that jointly models deterministic, latent, and abrupt-fluctuation dynamics within time series data. This design is not just a structural choice — it is a necessity: deterministic components provide a stable backbone, latent modeling enriches predictive expressiveness, and fluctuation emphasis ensures sensitivity to critical events. Only through the integration of these three pathways can robust high-fidelity predictions be achieved under non-stationary conditions while simultaneously achieving significant computational cost savings.

The model processes an input sequence $\mathbf{X} \in \mathbb{R}^{B \times L \times D}$ to produce a prediction $\hat{\mathbf{X}} \in \mathbb{R}^{B \times L^{\prime} \times D}$ through the synergistic operation of three specialized components, as illustrated in \textbf{Figure 2}. The forward process is a fusion of predictions from three parallel pathways, formalized by a residual connection that aggregates their outputs:

\begin{equation}
\hat{\mathbf{X}}=\underbrace{\hat{\mathbf{X}}^{(x)}}_{\text {Source Data }}+\underbrace{\hat{\mathbf{X}}^{(z)}}_{\text {Latent Information }}+\underbrace{\Delta_{\text {emphasis }}}_{\text {Volatility Information }}.
\end{equation}

\textbf{Trend Modeling $(\hat{\mathbf{X}}^{(x)})$.} This branch provides a stable, base-level forecast by applying series decomposition and linear projection directly on the input $\mathbf{X}$. It reliably captures broad, predictable patterns and trends, forming the foundation of the final prediction. To capture the latest sudden fluctuations, TimeCatcher incorporates the concept proposed by xPatch \cite{stitsyuk2025xpatch}, employing Exponential Moving Average seasonal-trend decomposition method instead of simple average pooling. This approach assigns higher weights to new data points in the input sequence:
\begin{equation}
    \mathbf{X}^{(s)}, \mathbf{X}^{(t)}=\mathcal{D}_{x}(\mathbf{X}), 
\end{equation}
where $\mathbf{X}^{(s)}$ and $\mathbf{X}^{(t)}$ represent the seasonal and trend components, respectively. TimeCatcher builds upon the ideas of Zeng et al. \cite{zeng2023transformers}, adopting an MLP architecture similar to DLinear. By simultaneously projecting both components into the prediction time domain through learnable linear layers, this approach significantly reduces training costs while achieving prediction performance comparable to that of the Transformer framework:
\begin{equation}
    \hat{\mathbf{X}}^{(x)}=W_{s}^{(x)} \mathbf{X}^{(s)}+W_{t}^{(x)} \mathbf{X}^{(t)}.
\end{equation}

\textbf{Latent Dynamic Encoding $(\hat{\mathbf{X}}^{(z)})$.}  This component employs a VAE to map the input into a stochastic latent space, forecasts the evolution of the latent states, and decodes them back to the data space. It is designed to catch complex, non-linear, and stochastic patterns that are not easily discernible from the source data alone. 

\textbf{Volatility Enhancement $(\Delta_{\text {emphasis }})$.} This module dynamically identifies significant transient fluctuations and abrupt fluctuations in the forecasted trajectory. It calculates an emphasis signal that amplifies these critical events, ensuring they are not smoothed over in the final output.

The final prediction $\hat{\mathbf{X}}$ is a simple yet powerful summation of these three elements. This residual design allows the model to confidently learn the base signal while adaptively refining it with more complex and transient patterns caught by the specialized VAE and Volatility Emphasis modules.

\subsection{Latent Dynamic Encoding}
Even with careful decomposition, a large portion of temporal dependency remains implicit — driven by stochastic patterns, inter-variable interactions, and unobserved exogenous influences. Modeling these dynamics directly from raw data is extremely difficult, often requiring massive models with poor interpretability.

To address this, TimeCatcher leverages a variational autoencoder (VAE) to project the input sequence into a structured latent space. This step is not optional — it is essential for three reasons:
\begin{itemize}
\item \textbf{Necessity for abstraction:} Latent variables z encode complex temporal structures beyond surface-level correlations.
\item \textbf{Probabilistic regularization:} The KL-regularized latent space ensures smoothness and continuity, improving generalization and reducing sensitivity to noise.
\item \textbf{Dynamic forecasting capability:} By evolving z into the future, we capture transitions that raw space models miss.
\end{itemize}

To project sequences into latent space, the \textbf{VAE encoder} first performs feature extraction and downsampling on the source data through N layers of one-dimensional convolutional layers with ReLU activation functions. The powerful local modeling capability and translation invariance of one-dimensional convolutional layers enable them to accurately capture local information in time series while requiring fewer training parameters than other architectures. After flattening, linear projections are applied to obtain the latent space mean vector $\boldsymbol{\mu}_{z}$ and the latent space log variance vector $\boldsymbol{\sigma}_{z}$.

Given input $\mathbf{X}$ and the encoder network $f_{\mathrm{enc}}$, the encoder parameterizes a Gaussian posterior distribution:
\begin{equation}
    q_{\phi}(\mathbf{z} \mid \mathbf{X})=\mathcal{N}\left(\mathbf{z} ; \boldsymbol{\mu}_{z}, \operatorname{diag}\left(\boldsymbol{\sigma}_{z}^{2}\right)\right),
\end{equation}
where
\begin{equation}
    \boldsymbol{\mu}_{z}, \log \boldsymbol{\sigma}_{z}^{2}=f_{\mathrm{enc}}(\mathbf{X} ; \phi).
\end{equation}
The latent variable $\mathbf{z} \in \mathbb{R}^{B \times H}$ is then sampled via the reparameterization trick:
\begin{equation}
    \mathbf{z}=\boldsymbol{\mu}_{z}+\boldsymbol{\sigma}_{z} \odot \boldsymbol{\epsilon}, \quad \boldsymbol{\epsilon} \sim \mathcal{N}(0, \mathbf{I}).
\end{equation}

Rather than using the sampled latent state $\mathbf{z}$ directly, we forecast its future evolution, a process grounded in the probabilistic framework of the VAE. The latent state $\mathbf{z}$ is a random variable drawn from a learned approximate posterior $q_{\phi}(\mathbf{z} \mid \mathbf{X})$, which is regularized towards a smooth prior $p(\mathbf{z}) \sim \mathcal{N}(\mathbf{0}, \mathbf{I})$ via the Kullback-Leibler (KL) divergence constraint in the evidence lower bound (ELBO) objective \cite{kingma2013auto}. This regularization imposes critical mathematical properties on the latent space: it encourages smoothness and continuity, meaning the probability density varies gradually and similar inputs map to nearby latent points, resulting in a more stable and structured temporal evolution of $\mathbf{z}$ compared to the raw observation space. Recognizing that $\mathbf{z}$ itself forms a meaningful and regularized time series, we decompose it into interpretable seasonal and trend components:
\begin{equation}
    \mathbf{z}^{(s)}, \mathbf{z}^{(t)}=\mathcal{D}_{z}(\mathbf{z}),
\end{equation}
and future latent states are predicted as:
\begin{equation}
    \hat{\mathbf{z}}=W_{s}^{(z)} \mathbf{z}^{(s)}+W_{t}^{(z)} \mathbf{z}^{(t)}.
\end{equation}
The weight matrices $W^{(z)}$ are initialized to promote a smooth transition.

Finally, for the decoder, we utilize a MLP to reconstruct the time-domain signal from the latent vector. This asymmetric design balances expressiveness and computational efficiency while still achieving satisfactory reconstruction quality. Through N stacked linear layers, the decoder network $f_{\mathrm{dec}}$ transforms the predicted latent state $\hat{\mathbf{z}}$ back into the data space, producing the latent information component of the forecast:
\begin{equation}
    \hat{\mathbf{X}}^{(z)}=f_{\mathrm{dec}}(\hat{\mathbf{z}} ; \theta).
\end{equation}

This latent-state pathway captures non-linear, high-order temporal dynamics that are difficult to model directly from raw data, providing a complementary representation to the deterministic forecasting branch.

\subsection{Volatility Enhancement}
Real-world time series often contain abrupt structural changes — such as regime shifts or anomalies — that conventional models tend to smooth out. To address this, TimeCatcher introduces a volatility enhancement module that explicitly enhances model responses to such events. Critical short-term fluctuations and change points often contain vital information. The Volatility Enhancement Module is designed to explicitly catch and amplify these events.

However, not all fluctuation information is learnable or predictable, such as minor data jitter caused by instrument measurements. To intelligently identify noteworthy abrupt changes, we designed a \textbf{dynamic time-aware mask}. Through sliding window statistics and a dynamic threshold mechanism, it adaptively identifies and generates a binary mask to highlight significant abrupt fluctuation regions in the sequence, providing precise localization for subsequent adaptive reinforcement. Specifically, for the latent-space reconstruction $\hat{\mathbf{X}}^{(z)}$, we calculate the local amplitude and directional change within a sliding window of size $w$:
\begin{equation}
    \mathbf{A}_{t}=\max \left(\hat{\mathbf{X}}_{t-w: t}^{(z)}\right)-\min \left(\hat{\mathbf{X}}_{t-w: t}^{(z)}\right),
\end{equation}
\begin{equation}
    \mathbf{D}_{t}=\operatorname{sign}\left(\hat{\mathbf{X}}_{t}^{(z)}-\hat{\mathbf{X}}_{t-w}^{(z)}\right).
\end{equation}
For each variate $d$ at $t$, the amplitude of fluctuation $\mathbf{A}_{t,d}$ within a sliding window is compared to a dynamic threshold $\tau$ derived from the input data's range:
\begin{equation}
    \operatorname{Mask}_{t, d}=\mathbb{I}\left(A_{t, d} \geq \tau \cdot\left(\max \left(\mathbf{X}_{:, d}\right)-\min \left(\mathbf{X}_{:, d}\right)\right)\right).
\end{equation}
This completes the temporal perception mask, identifying the volatility information that should be noted and studied.

Then, to highlight the difference regions and quantify the magnitude of the difference, the module computes the raw residual signal $\Delta_{\text {raw }}$ by comparing the VAE prediction with the baseline prediction:
\begin{equation}
    \Delta_{\text {raw }}=\left|\hat{\mathbf{X}}_{t}^{(z)}-\hat{\mathbf{X}}_{t}^{(x)}\right|.
\end{equation}
\begin{table*}[h]
\centering
\caption{Full results on the long-term prediction task. We compare against a wide range of competing models at different prediction lengths. The average is the mean over all four prediction lengths, i.e., $\{96, 192, 336, 720\}$.}
\label{tab:performance_comparison}
\resizebox{\textwidth}{!}{
\begin{tabular}{cc|cc|cc|cc|cc|cc|cc|cc|cc|cc|cc|cc|cc}
\toprule
\multicolumn{2}{c}{\multirow{2}{*}{\textbf{Models}}}
 & \multicolumn{2}{c|}{\textbf{TimeCatcher}} & \multicolumn{2}{c|}{\textbf{TimeBase}} & \multicolumn{2}{c|}{\textbf{TimeMixer}} & \multicolumn{2}{c|}{\textbf{iTransformer}} & \multicolumn{2}{c|}{\textbf{PatchTST}} & \multicolumn{2}{c|}{\textbf{Crossformer}} & \multicolumn{2}{c|}{\textbf{TiDE}} & \multicolumn{2}{c|}{\textbf{TimesNet}} & \multicolumn{2}{c|}{\textbf{DLinear}} & \multicolumn{2}{c|}{\textbf{SCINet}} & \multicolumn{2}{c|}{\textbf{FEDformer}} & \multicolumn{2}{c}{\textbf{Autoformer}} \\
\multicolumn{2}{c}{} & \multicolumn{2}{c|}{\textbf{(Ours)}} & \multicolumn{2}{c|}{\textbf{(2025)}} & \multicolumn{2}{c|}{\textbf{(2024b)}} & \multicolumn{2}{c|}{\textbf{(2024)}} & \multicolumn{2}{c|}{\textbf{(2023)}} & \multicolumn{2}{c|}{\textbf{(2023)}} & \multicolumn{2}{c|}{\textbf{(2023a)}} & \multicolumn{2}{c|}{\textbf{(2023)}} & \multicolumn{2}{c|}{\textbf{(2023)}} & \multicolumn{2}{c|}{\textbf{(2022a)}} & \multicolumn{2}{c|}{\textbf{(2022b)}} & \multicolumn{2}{c}{\textbf{(2021)}} \\
\cmidrule(lr){3-4}\cmidrule(lr){5-6}\cmidrule(lr){7-8}\cmidrule(lr){9-10}\cmidrule(lr){11-12}\cmidrule(lr){13-14}\cmidrule(lr){15-16}\cmidrule(lr){17-18}\cmidrule(lr){19-20}\cmidrule(lr){21-22}\cmidrule(lr){23-24}\cmidrule(lr){25-26}
\multicolumn{2}{c}{\textbf{Metric}} & \textbf{MSE} & \textbf{MAE} & \textbf{MSE} & \textbf{MAE} & \textbf{MSE} & \textbf{MAE} & \textbf{MSE} & \textbf{MAE} & \textbf{MSE} & \textbf{MAE} & \textbf{MSE} & \textbf{MAE} & \textbf{MSE} & \textbf{MAE} & \textbf{MSE} & \textbf{MAE} & \textbf{MSE} & \textbf{MAE} & \textbf{MSE} & \textbf{MAE} & \textbf{MSE} & \textbf{MAE} & \textbf{MSE} & \textbf{MAE} \\
\midrule
\multicolumn{1}{c|}{\multirow{5}{*}{\rotatebox[origin=c]{90}{Electricity}}} & 96 & \textcolor{red}{\textbf{0.147}} & \textcolor{red}{\textbf{0.238}} & 0.212 & 0.278 & 0.153 & 0.247 & \textcolor{blue}{0.148} & \textcolor{blue}{0.240} & 0.190 & 0.296 & 0.219 & 0.314 & 0.237 & 0.329 & 0.168 & 0.272 & 0.201 & 0.302 & 0.247 & 0.345 & 0.193 & 0.308 & 0.201 & 0.317 \\
 \multicolumn{1}{c|}{} & 192 & \textcolor{red}{\textbf{0.154}} & \textcolor{red}{\textbf{0.252}} & 0.209 & 0.280 & 0.166 & 0.256 & \textcolor{blue}{0.162} & \textcolor{blue}{0.253} & 0.199 & 0.304 & 0.231 & 0.322 & 0.236 & 0.330 & 0.184 & 0.322 & 0.210 & 0.305 & 0.257 & 0.355 & 0.201 & 0.315 & 0.222 & 0.334 \\
 \multicolumn{1}{c|}{} & 336 & \textcolor{red}{\textbf{0.163}} & \textcolor{red}{\textbf{0.261}} & 0.222 & 0.294 & 0.185 & 0.277 & \textcolor{blue}{0.178} & \textcolor{blue}{0.269} & 0.217 & 0.319 & 0.246 & 0.337 & 0.249 & 0.344 & 0.198 & 0.300 & 0.223 & 0.319 & 0.269 & 0.369 & 0.214 & 0.329 & 0.231 & 0.443 \\
 \multicolumn{1}{c|}{} & 720 & \textcolor{red}{\textbf{0.208}} & \textcolor{red}{\textbf{0.309}} & 0.264 & 0.327 & 0.225 & \textcolor{blue}{0.310} & 0.225 & 0.317 & 0.258 & 0.352 & 0.280 & 0.363 & 0.284 & 0.373 & \textcolor{blue}{0.220} & 0.320 & 0.258 & 0.350 & 0.299 & 0.390 & 0.246 & 0.355 & 0.254 & 0.361 \\
 \cmidrule(lr){2-2}\cmidrule(lr){3-4}\cmidrule(lr){5-6}\cmidrule(lr){7-8}\cmidrule(lr){9-10}\cmidrule(lr){11-12}\cmidrule(lr){13-14}\cmidrule(lr){15-16}\cmidrule(lr){17-18}\cmidrule(lr){19-20}\cmidrule(lr){21-22}\cmidrule(lr){23-24}\cmidrule(lr){25-26}
 \multicolumn{1}{c|}{} & Avg & \textcolor{red}{\textbf{0.168}} & \textcolor{red}{\textbf{0.265}} & 0.227 & 0.295 & 0.182 & 0.272 & \textcolor{blue}{0.178} & \textcolor{blue}{0.270} & 0.205 & 0.290 & 0.244 & 0.334 & 0.251 & 0.344 & 0.192 & 0.295 & 0.212 & 0.300 & 0.268 & 0.365 & 0.214 & 0.327 & 0.227 & 0.338 \\
 \midrule
 \multicolumn{1}{c|}{\multirow{5}{*}{\rotatebox[origin=c]{90}{Exchange}}} & 96 & {\color[HTML]{FF0000} \textbf{0.080}} & {\color[HTML]{FF0000} \textbf{0.199}} & 0.103 & 0.228 & 0.090 & 0.235 & \textcolor{blue}{0.086} & 0.206                                 & 0.088                                 & \textcolor{blue}{0.205} & 0.256 & 0.367 & 0.094 & 0.218 & 0.107 & 0.234 & 0.088                                 & 0.218 & 0.267 & 0.396 & 0.148 & 0.278 & 0.197 & 0.323 \\
 \multicolumn{1}{c|}{} & 192 & {\color[HTML]{FF0000} \textbf{0.169}} & \textcolor{blue}{0.306} & 0.194 & 0.317 & 0.187 & 0.343 & 0.177                                 & {\color[HTML]{FF0000} \textbf{0.299}} & \textcolor{blue}{0.176} & {\color[HTML]{FF0000} \textbf{0.299}} & 0.470 & 0.509 & 0.184 & 0.307 & 0.226 & 0.344 & 0.176                                 & 0.315 & 0.351 & 0.459 & 0.271 & 0.315 & 0.300 & 0.369 \\
 \multicolumn{1}{c|}{} & 336 & {\color[HTML]{FF0000} \textbf{0.296}} & \textcolor{blue}{0.415} & 0.353 & 0.431 & 0.353 & 0.473 & 0.331                                 & 0.417                                 & \textcolor{blue}{0.301} & {\color[HTML]{FF0000} \textbf{0.397}} & 1.268 & 0.883 & 0.349 & 0.431 & 0.367 & 0.448 & 0.313                                 & 0.427 & 1.324 & 0.853 & 0.460 & 0.427 & 0.509 & 0.524 \\
 \multicolumn{1}{c|}{} & 720 & {\color[HTML]{FF0000} \textbf{0.749}} & {\color[HTML]{FF0000} \textbf{0.652}} & 1.158 & 0.830 & 0.934 & 0.761 & 0.847                                 & \textcolor{blue}{0.691} & 0.901                                 & 0.714                                 & 1.767 & 1.068 & 0.852 & 0.698 & 0.964 & 0.746 & \textcolor{blue}{0.839} & 0.695 & 1.058 & 0.797 & 1.195 & 0.695 & 1.447 & 0.941 \\
 \cmidrule(lr){2-2}\cmidrule(lr){3-4}\cmidrule(lr){5-6}\cmidrule(lr){7-8}\cmidrule(lr){9-10}\cmidrule(lr){11-12}\cmidrule(lr){13-14}\cmidrule(lr){15-16}\cmidrule(lr){17-18}\cmidrule(lr){19-20}\cmidrule(lr){21-22}\cmidrule(lr){23-24}\cmidrule(lr){25-26}
 \multicolumn{1}{c|}{} & Avg & {\color[HTML]{FF0000} \textbf{0.324}} & {\color[HTML]{FF0000} \textbf{0.393}} & 0.452 & 0.452 & 0.391 & 0.453 & 0.360                                 & \textcolor{blue}{0.403} & 0.367                                 & 0.404                                 & 0.940 & 0.707 & 0.370 & 0.414 & 0.416 & 0.443 & \textcolor{blue}{0.354} & 0.414 & 0.750 & 0.626 & 0.519 & 0.429 & 0.613 & 0.539 \\
 \midrule
 \multicolumn{1}{c|}{\multirow{5}{*}{\rotatebox[origin=c]{90}{Solar Energy}}} & 96 & {\color[HTML]{FF0000} \textbf{0.182}} & {\color[HTML]{FF0000} \textbf{0.200}} & 0.673 & 0.445 & \textcolor{blue}{0.189} & 0.259 & 0.203 & \textcolor{blue}{0.237} & 0.265 & 0.323 & 0.232 & 0.302 & 0.312 & 0.399 & 0.373 & 0.358 & 0.290 & 0.378 & 0.237 & 0.344 & 0.286 & 0.341 & 0.456 & 0.446 \\
 \multicolumn{1}{c|}{} & 192 & {\color[HTML]{FF0000} \textbf{0.214}} & {\color[HTML]{FF0000} \textbf{0.224}} & 0.692 & 0.454 & \textcolor{blue}{0.222} & 0.283 & 0.233 & \textcolor{blue}{0.261} & 0.288 & 0.332 & 0.371 & 0.410 & 0.339 & 0.416 & 0.397 & 0.376 & 0.320 & 0.398 & 0.280 & 0.380 & 0.291 & 0.337 & 0.588 & 0.561 \\
 \multicolumn{1}{c|}{} & 336 & {\color[HTML]{FF0000} \textbf{0.220}} & {\color[HTML]{FF0000} \textbf{0.233}} & 0.682 & 0.444 & \textcolor{blue}{0.231} & 0.292 & 0.248 & \textcolor{blue}{0.273} & 0.301 & 0.339 & 0.495 & 0.515 & 0.368 & 0.430 & 0.420 & 0.380 & 0.353 & 0.415 & 0.304 & 0.389 & 0.354 & 0.416 & 0.595 & 0.588 \\
 \multicolumn{1}{c|}{} & 720 & {\color[HTML]{FF0000} \textbf{0.216}} & {\color[HTML]{FF0000} \textbf{0.231}} & 0.625 & 0.411 & \textcolor{blue}{0.223} & 0.285 & 0.249 & \textcolor{blue}{0.275} & 0.295 & 0.336 & 0.526 & 0.542 & 0.370 & 0.425 & 0.420 & 0.381 & 0.357 & 0.413 & 0.308 & 0.388 & 0.380 & 0.437 & 0.733 & 0.633 \\
 \cmidrule(lr){2-2}\cmidrule(lr){3-4}\cmidrule(lr){5-6}\cmidrule(lr){7-8}\cmidrule(lr){9-10}\cmidrule(lr){11-12}\cmidrule(lr){13-14}\cmidrule(lr){15-16}\cmidrule(lr){17-18}\cmidrule(lr){19-20}\cmidrule(lr){21-22}\cmidrule(lr){23-24}\cmidrule(lr){25-26}
 \multicolumn{1}{c|}{} & Avg & {\color[HTML]{FF0000} \textbf{0.208}} & {\color[HTML]{FF0000} \textbf{0.222}} & 0.668 & 0.439 & \textcolor{blue}{0.216} & 0.280 & 0.233 & \textcolor{blue}{0.262} & 0.287 & 0.333 & 0.406 & 0.442 & 0.347 & 0.418 & 0.403 & 0.374 & 0.330 & 0.401 & 0.282 & 0.375 & 0.328 & 0.383 & 0.593 & 0.557 \\
 \midrule
 \multicolumn{1}{c|}{\multirow{5}{*}{\rotatebox[origin=c]{90}{ETTh1}}} & 96 & \textcolor{blue}{0.382} & {\color[HTML]{FF0000} \textbf{0.389}} & 0.398 & \textcolor{blue}{0.391} & {\color[HTML]{FF0000} \textbf{0.375}} & 0.400  & 0.386 & 0.405 & 0.460 & 0.447 & 0.423 & 0.448 & 0.479 & 0.464 & 0.384  & 0.402 & 0.397 & 0.412 & 0.654 & 0.599 & 0.395 & 0.424 & 0.449 & 0.459 \\
 \multicolumn{1}{c|}{} & 192 & 0.471                                 & 0.428                                 & 0.455 & \textcolor{blue}{0.423} & {\color[HTML]{FF0000} \textbf{0.429}} & {\color[HTML]{FF0000} \textbf{0.421}} & 0.441 & 0.512 & 0.477 & 0.429 & 0.471 & 0.474 & 0.525 & 0.492 & \textcolor{blue}{0.436} & 0.429 & 0.446 & 0.441 & 0.719 & 0.631 & 0.469 & 0.470 & 0.500 & 0.482 \\
 \multicolumn{1}{c|}{} & 336 & {\color[HTML]{FF0000} \textbf{0.443}} & {\color[HTML]{FF0000} \textbf{0.426}} & 0.501 & \textcolor{blue}{0.443} & \textcolor{blue}{0.484} & 0.458                                 & 0.487 & 0.458 & 0.546 & 0.496 & 0.570 & 0.546 & 0.565 & 0.515 & 0.491                                 & 0.469 & 0.489 & 0.467 & 0.778 & 0.659 & 0.530 & 0.499 & 0.521 & 0.496 \\
 \multicolumn{1}{c|}{} & 720 & {\color[HTML]{FF0000} \textbf{0.476}} & \textcolor{blue}{0.472} & \textcolor{blue}{0.498} & {\color[HTML]{FF0000} \textbf{0.458}} & \textcolor{blue}{0.498}                                 & 0.482                                 & 0.503 & 0.491 & 0.544 & 0.517 & 0.653 & 0.621 & 0.594 & 0.558 & 0.521                                 & 0.500 & 0.513 & 0.510 & 0.836 & 0.699 & 0.598 & 0.544 & 0.514 & 0.512 \\
 \cmidrule(lr){2-2}\cmidrule(lr){3-4}\cmidrule(lr){5-6}\cmidrule(lr){7-8}\cmidrule(lr){9-10}\cmidrule(lr){11-12}\cmidrule(lr){13-14}\cmidrule(lr){15-16}\cmidrule(lr){17-18}\cmidrule(lr){19-20}\cmidrule(lr){21-22}\cmidrule(lr){23-24}\cmidrule(lr){25-26}
 \multicolumn{1}{c|}{} & Avg & {\color[HTML]{FF0000} \textbf{0.443}} & {\color[HTML]{FF0000} \textbf{0.429}} & 0.463 & {\color[HTML]{FF0000} \textbf{0.429}} &  \textcolor{blue}{0.447}                                & \textcolor{blue}{0.440}                                 & 0.454 & 0.467 & 0.507 & 0.472 & 0.529 & 0.522 & 0.541 & 0.507 & 0.458                                 & 0.450 & 0.461 & 0.458 & 0.747 & 0.647 & 0.498 & 0.484 & 0.496 & 0.487 \\
 \midrule
 \multicolumn{1}{c|}{\multirow{5}{*}{\rotatebox[origin=c]{90}{ETTh2}}} & 96 & {\color[HTML]{FF0000} \textbf{0.271}} & \textcolor{blue}{0.348}   & 0.337 & 0.376 & \textcolor{blue}{0.289}   & {\color[HTML]{FF0000} \textbf{0.341}} & 0.297                                 & 0.349 & 0.297 & 0.349 & 0.745 & 0.584 & 0.400 & 0.440 & 0.340 & 0.374 & 0.340 & 0.394 & 0.707 & 0.621 & 0.358 & 0.397 & 0.346 & 0.388 \\
 \multicolumn{1}{c|}{} & 192 & {\color[HTML]{FF0000} \textbf{0.338}} & {\color[HTML]{FF0000} \textbf{0.390}} & 0.401 & 0.405 & \textcolor{blue}{0.372} & \textcolor{blue}{0.392} & 0.380                                 & 0.400                                 & 0.380 & 0.400                                 & 0.877 & 0.656 & 0.528 & 0.509 & 0.402 & 0.414 & 0.482 & 0.479 & 0.860 & 0.689 & 0.429 & 0.439 & 0.456 & 0.452 \\
 \multicolumn{1}{c|}{} & 336 & {\color[HTML]{FF0000} \textbf{0.377}} & \textcolor{blue}{0.419} & 0.436 & 0.442 & \textcolor{blue}{0.386}                                 & \textcolor{red}{\textbf{0.414}} & 0.428                                 & 0.432                                 & 0.428 & 0.432                                 & 1.043 & 0.731 & 0.643 & 0.571 & 0.452 & 0.452 & 0.591 & 0.541 & 1.000 & 0.744 & 0.496 & 0.487 & 0.482 & 0.486 \\
 \multicolumn{1}{c|}{} & 720 & 0.442                                 & 0.456                                 & 0.459 & 0.477 & {\color[HTML]{FF0000} \textbf{0.412}} & {\color[HTML]{FF0000} \textbf{0.434}} & \textcolor{blue}{0.427} & \textcolor{blue}{0.445} & 0.436 & 0.450                                 & 1.104 & 0.763 & 0.874 & 0.679 & 0.462 & 0.468 & 0.839 & 0.661 & 1.249 & 0.838 & 0.463 & 0.474 & 0.515 & 0.511 \\
 \cmidrule(lr){2-2}\cmidrule(lr){3-4}\cmidrule(lr){5-6}\cmidrule(lr){7-8}\cmidrule(lr){9-10}\cmidrule(lr){11-12}\cmidrule(lr){13-14}\cmidrule(lr){15-16}\cmidrule(lr){17-18}\cmidrule(lr){19-20}\cmidrule(lr){21-22}\cmidrule(lr){23-24}\cmidrule(lr){25-26}
 \multicolumn{1}{c|}{} & Avg & {\color[HTML]{FF0000} \textbf{0.357}} & \textcolor{blue}{0.404} & 0.408 & 0.425 & \textcolor{blue}{0.365} & {\color[HTML]{FF0000} \textbf{0.395}} & 0.383                                 & 0.407                                 & 0.385 & 0.408                                 & 0.942 & 0.684 & 0.611 & 0.550 & 0.414 & 0.427 & 0.563 & 0.519 & 0.954 & 0.723 & 0.437 & 0.449 & 0.450 & 0.459 \\
 \midrule
 \multicolumn{1}{c|}{\multirow{5}{*}{\rotatebox[origin=c]{90}{ETTm1}}} & 96 & \textcolor{blue}{0.329} & {\color[HTML]{FF0000} \textbf{0.356}} & 0.373 & 0.387 & {\color[HTML]{FF0000} \textbf{0.320}} & \textcolor{blue}{0.357} & 0.334 & 0.368 & 0.334 & 0.368 & 0.404 & 0.426 & 0.364 & 0.387 & 0.338 & 0.375 & 0.346 & 0.374 & 0.418 & 0.438 & 0.379 & 0.419 & 0.505 & 0.475 \\
 \multicolumn{1}{c|}{} & 192 & {\color[HTML]{FF0000} \textbf{0.356}} & {\color[HTML]{FF0000} \textbf{0.373}} & 0.410 & 0.408 & \textcolor{blue}{0.361} & \textcolor{blue}{0.381} & 0.390 & 0.393 & 0.390 & 0.393 & 0.450 & 0.451 & 0.398 & 0.404 & 0.374 & 0.387 & 0.382 & 0.391 & 0.439 & 0.450 & 0.426 & 0.441 & 0.553 & 0.496 \\
 \multicolumn{1}{c|}{} & 336 & \textcolor{blue}{0.395} & {\color[HTML]{FF0000} \textbf{0.396}} & 0.435 & 0.420 & {\color[HTML]{FF0000} \textbf{0.390}} & \textcolor{blue}{0.404} & 0.426 & 0.420 & 0.426 & 0.420 & 0.532 & 0.515 & 0.428 & 0.425 & 0.410 & 0.411 & 0.415 & 0.415 & 0.490 & 0.485 & 0.445 & 0.459 & 0.621 & 0.537 \\
 \multicolumn{1}{c|}{} & 720 & \textcolor{blue}{0.467} & {\color[HTML]{FF0000} \textbf{0.439}} & 0.503 & 0.461 & {\color[HTML]{FF0000} \textbf{0.454}} & \textcolor{blue}{0.441} & 0.491 & 0.459 & 0.491 & 0.459 & 0.666 & 0.589 & 0.487 & 0.461 & 0.478 & 0.450 & 0.473 & 0.451 & 0.595 & 0.550 & 0.543 & 0.490 & 0.671 & 0.561 \\
 \cmidrule(lr){2-2}\cmidrule(lr){3-4}\cmidrule(lr){5-6}\cmidrule(lr){7-8}\cmidrule(lr){9-10}\cmidrule(lr){11-12}\cmidrule(lr){13-14}\cmidrule(lr){15-16}\cmidrule(lr){17-18}\cmidrule(lr){19-20}\cmidrule(lr){21-22}\cmidrule(lr){23-24}\cmidrule(lr){25-26}
 \multicolumn{1}{c|}{} & Avg & \textcolor{blue}{0.387} & {\color[HTML]{FF0000} \textbf{0.391}} & 0.430 & 0.419 & {\color[HTML]{FF0000} \textbf{0.381}} & \textcolor{blue}{0.396} & 0.410 & 0.410 & 0.410 & 0.410 & 0.513 & 0.495 & 0.419 & 0.419 & 0.400 & 0.406 & 0.404 & 0.408 & 0.486 & 0.481 & 0.448 & 0.452 & 0.588 & 0.517 \\
 \midrule
 \multicolumn{1}{c|}{\multirow{5}{*}{\rotatebox[origin=c]{90}{ETTm2}}} & 96 & {\color[HTML]{FF0000} \textbf{0.171}} & {\color[HTML]{FF0000} \textbf{0.258}} & 0.187                                 & 0.270                                 & \textcolor{blue}{0.175} & {\color[HTML]{FF0000} \textbf{0.258}} & 0.180 & \textcolor{blue}{0.264} & 0.183 & 0.270 & 0.287 & 0.366 & 0.207 & 0.305 & 0.187                                 & 0.267 & 0.193 & 0.293 & 0.286 & 0.377 & 0.203 & 0.287 & 0.255 & 0.339 \\
 \multicolumn{1}{c|}{} & 192 & {\color[HTML]{FF0000} \textbf{0.217}} & {\color[HTML]{FF0000} \textbf{0.292}} & 0.250                                 & 0.308                                 & \textcolor{blue}{0.237} & \textcolor{blue}{0.299} & 0.250 & 0.309                                 & 0.250 & 0.309 & 0.414 & 0.492 & 0.290 & 0.364 & 0.249                                 & 0.309 & 0.284 & 0.361 & 0.399 & 0.445 & 0.269 & 0.328 & 0.281 & 0.340 \\
 \multicolumn{1}{c|}{} & 336 & \textcolor{blue}{0.303} & 0.354                                 & 0.310                                 & \textcolor{blue}{0.345} & {\color[HTML]{FF0000} \textbf{0.298}} & {\color[HTML]{FF0000} \textbf{0.340}} & 0.311 & 0.348                                 & 0.311 & 0.348 & 0.597 & 0.542 & 0.377 & 0.422 & 0.321                                 & 0.351 & 0.382 & 0.429 & 0.637 & 0.591 & 0.325 & 0.366 & 0.339 & 0.372 \\
 \multicolumn{1}{c|}{} & 720 & \textcolor{blue}{0.408} & 0.421                                 & 0.411                                 & \textcolor{blue}{0.401} & {\color[HTML]{FF0000} \textbf{0.391}} & {\color[HTML]{FF0000} \textbf{0.396}} & 0.412 & 0.407                                 & 0.412 & 0.407 & 1.730 & 1.042 & 0.558 & 0.524 & \textcolor{blue}{0.408} & 0.403 & 0.558 & 0.525 & 0.960 & 0.735 & 0.421 & 0.415 & 0.433 & 0.432 \\
 \cmidrule(lr){2-2}\cmidrule(lr){3-4}\cmidrule(lr){5-6}\cmidrule(lr){7-8}\cmidrule(lr){9-10}\cmidrule(lr){11-12}\cmidrule(lr){13-14}\cmidrule(lr){15-16}\cmidrule(lr){17-18}\cmidrule(lr){19-20}\cmidrule(lr){21-22}\cmidrule(lr){23-24}\cmidrule(lr){25-26}
 \multicolumn{1}{c|}{} & Avg & {\color[HTML]{FF0000} \textbf{0.275}} & \textcolor{blue}{0.331} & \textcolor{blue}{0.290} & \textcolor{blue}{0.331} & {\color[HTML]{FF0000} \textbf{0.275}} & {\color[HTML]{FF0000} \textbf{0.323}} & 0.288 & 0.332                                 & 0.289 & 0.334 & 0.757 & 0.611 & 0.358 & 0.404 & 0.291                                 & 0.333 & 0.354 & 0.402 & 0.571 & 0.537 & 0.305 & 0.349 & 0.327 & 0.371 \\
 \midrule
 \multicolumn{1}{c|}{\multirow{5}{*}{\rotatebox[origin=c]{90}{Weather}}} & 96 & \textcolor{blue}{0.166} & 0.217                                 & 0.169                                 & 0.215 & {\color[HTML]{FF0000} \textbf{0.163}} & {\color[HTML]{FF0000} \textbf{0.209}} & 0.174 & \textcolor{blue}{0.214} & 0.186 & 0.227 & 0.195                                 & 0.271 & 0.202 & 0.261 & 0.172 & 0.220 & 0.195 & 0.252 & 0.221 & 0.306 & 0.217 & 0.296 & 0.266 & 0.336 \\
 \multicolumn{1}{c|}{} & 192 & 0.212                                 & \textcolor{blue}{0.252} & 0.215                                 & 0.256 & {\color[HTML]{FF0000} \textbf{0.208}} & {\color[HTML]{FF0000} \textbf{0.250}} & 0.221 & 0.254                                 & 0.234 & 0.265 & \textcolor{blue}{0.209} & 0.277 & 0.242 & 0.298 & 0.219 & 0.261 & 0.237 & 0.295 & 0.261 & 0.340 & 0.276 & 0.336 & 0.307 & 0.367 \\
 \multicolumn{1}{c|}{} & 336 & {\color[HTML]{FF0000} \textbf{0.251}} & {\color[HTML]{FF0000} \textbf{0.285}} & \textcolor{blue}{0.271} & 0.296 & {\color[HTML]{FF0000} \textbf{0.251}} & \textcolor{blue}{0.287} & 0.278 & 0.296                                 & 0.284 & 0.301 & 0.273                                 & 0.332 & 0.287 & 0.335 & 0.280 & 0.306 & 0.282 & 0.331 & 0.309 & 0.378 & 0.339 & 0.380 & 0.359 & 0.395 \\
 \multicolumn{1}{c|}{} & 720 & {\color[HTML]{FF0000} \textbf{0.307}} & {\color[HTML]{FF0000} \textbf{0.334}} & 0.351                                 & 0.347 & \textcolor{blue}{0.339} & \textcolor{blue}{0.341} & 0.358 & 0.347                                 & 0.356 & 0.349 & 0.379                                 & 0.401 & 0.351 & 0.386 & 0.365 & 0.359 & 0.345 & 0.382 & 0.377 & 0.427 & 0.403 & 0.428 & 0.419 & 0.428 \\
 \cmidrule(lr){2-2}\cmidrule(lr){3-4}\cmidrule(lr){5-6}\cmidrule(lr){7-8}\cmidrule(lr){9-10}\cmidrule(lr){11-12}\cmidrule(lr){13-14}\cmidrule(lr){15-16}\cmidrule(lr){17-18}\cmidrule(lr){19-20}\cmidrule(lr){21-22}\cmidrule(lr){23-24}\cmidrule(lr){25-26}
 \multicolumn{1}{c|}{} & Avg & {\color[HTML]{FF0000} \textbf{0.234}} & {\color[HTML]{FF0000} \textbf{0.272}} & 0.252                                 & 0.279 & \textcolor{blue}{0.240} & {\color[HTML]{FF0000} \textbf{0.272}} & 0.258 & \textcolor{blue}{0.278} & 0.265 & 0.286 & 0.264                                 & 0.320 & 0.271 & 0.320 & 0.259 & 0.287 & 0.265 & 0.315 & 0.292 & 0.363 & 0.309 & 0.360 & 0.338 & 0.382 \\
 \midrule
 \multicolumn{1}{c|}{\multirow{5}{*}{\rotatebox[origin=c]{90}{Traffic}}} & 96 & 0.484 & \textcolor{blue}{0.275} & 0.712 & 0.384 & \textcolor{blue}{0.462} & 0.285                                 & {\color[HTML]{FF0000} \textbf{0.395}} & {\color[HTML]{FF0000} \textbf{0.268}} & 0.526 & 0.347 & 0.644 & 0.429 & 0.805 & 0.493 & 0.593 & 0.321 & 0.650 & 0.396 & 0.788 & 0.499 & 0.587 & 0.366 & 0.613 & 0.388 \\
 \multicolumn{1}{c|}{} & 192 & 0.485 & \textcolor{blue}{0.280} & 0.652 & 0.362 & \textcolor{blue}{0.473} & 0.296                                 & {\color[HTML]{FF0000} \textbf{0.417}} & {\color[HTML]{FF0000} \textbf{0.276}} & 0.522 & 0.332 & 0.665 & 0.431 & 0.756 & 0.474 & 0.617 & 0.336 & 0.598 & 0.370 & 0.789 & 0.604 & 0.604 & 0.373 & 0.616 & 0.382 \\
 \multicolumn{1}{c|}{} & 336 & 0.503 & \textcolor{blue}{0.292} & 0.659 & 0.365 & \textcolor{blue}{0.498} & 0.296                                 & {\color[HTML]{FF0000} \textbf{0.433}} & {\color[HTML]{FF0000} \textbf{0.283}} & 0.517 & 0.334 & 0.674 & 0.420 & 0.762 & 0.477 & 0.629 & 0.336 & 0.605 & 0.373 & 0.797 & 0.508 & 0.621 & 0.383 & 0.622 & 0.337 \\
 \multicolumn{1}{c|}{} & 720 & 0.560 & 0.322                                 & 0.702 & 0.386 & \textcolor{blue}{0.506} & \textcolor{blue}{0.313} & {\color[HTML]{FF0000} \textbf{0.467}} & {\color[HTML]{FF0000} \textbf{0.302}} & 0.552 & 0.352 & 0.683 & 0.424 & 0.719 & 0.449 & 0.640 & 0.350 & 0.645 & 0.394 & 0.841 & 0.523 & 0.626 & 0.382 & 0.660 & 0.408 \\
 \cmidrule(lr){2-2}\cmidrule(lr){3-4}\cmidrule(lr){5-6}\cmidrule(lr){7-8}\cmidrule(lr){9-10}\cmidrule(lr){11-12}\cmidrule(lr){13-14}\cmidrule(lr){15-16}\cmidrule(lr){17-18}\cmidrule(lr){19-20}\cmidrule(lr){21-22}\cmidrule(lr){23-24}\cmidrule(lr){25-26}
 \multicolumn{1}{c|}{} & Avg & 0.508 & \textcolor{blue}{0.292} & 0.681 & 0.374 & \textcolor{blue}{0.485} & 0.298                                 & {\color[HTML]{FF0000} \textbf{0.428}} & {\color[HTML]{FF0000} \textbf{0.282}} & 0.529 & 0.341 & 0.667 & 0.426 & 0.761 & 0.473 & 0.620 & 0.336 & 0.625 & 0.383 & 0.804 & 0.534 & 0.610 & 0.376 & 0.628 & 0.379 \\
  \midrule
   \multicolumn{2}{c|}{$1^{st}$ Count} & {\color[HTML]{FF0000} \textbf{28}} & {\color[HTML]{FF0000} \textbf{27}} & 0 & 2 & \textcolor{blue}{13} & \textcolor{blue}{12} & 5 & 6 & 0 & 2 & 0 & 0 & 0 & 0 & 0 & 0 & 0 & 0 & 0 & 0 & 0 & 0 & 0 & 0 \\
\bottomrule
\end{tabular}%
}
\end{table*}
To ensure a monotonic, positive, and interpretable relationship between residual magnitude and emphasis intensity, we employ a learnable parameter $\gamma$ for scaling and process it through a linear layer $g(\cdot)$ to generate the emphasis intensity $M_{\text {emphasis }}$. The SoftPlus activation function is adopted to maintain stability and robustness. This approach is not only concise and efficient but also avoids overreacting to noise.
\begin{equation}
    M_{\text {emphasis }}=g\left(\operatorname{SoftPlus}(\gamma) \cdot\left|\Delta_{\text {raw }}\right|\right).
\end{equation}

The final volatility emphasis output, which constitutes the third catch, is the product of the magnitude, the direction, and the mask:
\begin{equation}
    \Delta_{\text {emphasis }}=M_{\text {emphasis }} \odot \mathbf{D}_{t} \odot \text { Mask }.
\end{equation}

Through this mechanism, TimeCatcher dynamically amplifies informative structural variations, significantly improving forecasting robustness in volatile or event-driven environments.

\section{Experiments}
\textbf{Datasets.} Our experiments utilize nine widely adopted real-world multivariate time series datasets to evaluate the performance and efficiency of TimeCatcher on ten real-world multivariate time series benchmarks. The datasets include Traffic, ETT (including ETTh1, ETTh2, ETTm1, and ETTm2), Weather, Solar Energy, Exchange Rate, and Electricity, following the experimental settings in \cite{zhou2021informer, liu2022scinet}.

\textbf{Baselines.} We compare TimeCatcher against 11 benchmark models, including the advanced long-term forecasting model TimeMixer (2024) \cite{wang2024timemixer}, the similarly lightweight model TimeBase (2025) \cite{huangtimebase}, advanced short-term forecasting models, and other competitive models such as iTransformer (2024) \cite{liu2023itransformer}, PatchTST (2023) \cite{nie2022time}, Crossformer (2023) \cite{zhang2023crossformer}, TiDE (2023) \cite{das2023long}, TimesNet (2023a) \cite{wu2022timesnet}, DLinear (2023) \cite{zeng2023transformers}, SCINet (2022a) \cite{liu2022scinet}, FEDformer (2022b) \cite{zhou2022fedformer}, and Autoformer (2021) \cite{wu2021autoformer}.

\begin{figure*}[htbp]
  \centering
  \begin{subfigure}[b]{0.32\textwidth}
    \includegraphics[width=\linewidth]{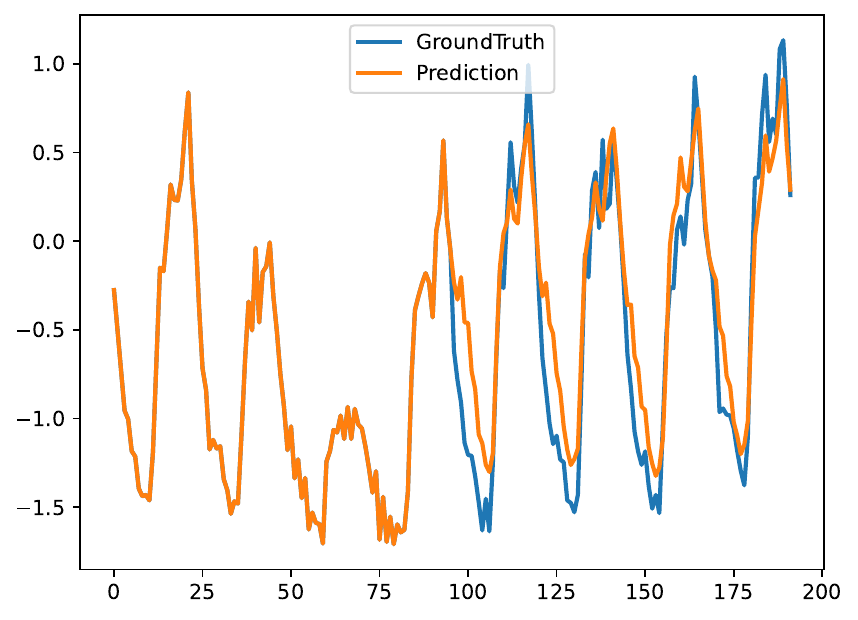}
    \caption{Prediction Visualization of TimeCatcher}
    \label{fig:temp}
  \end{subfigure}
  \hfill
  \begin{subfigure}[b]{0.32\textwidth}
    \includegraphics[width=\linewidth]{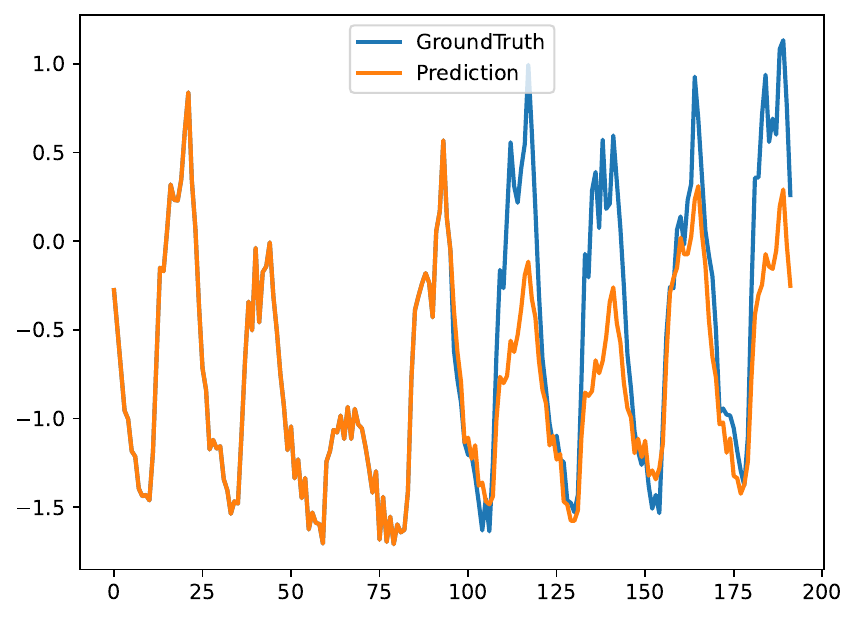}
    \caption{Prediction Visualization of TimeMixer}
    \label{fig:mem}
  \end{subfigure}
  \hfill
  \begin{subfigure}[b]{0.32\textwidth}
    \includegraphics[width=\linewidth]{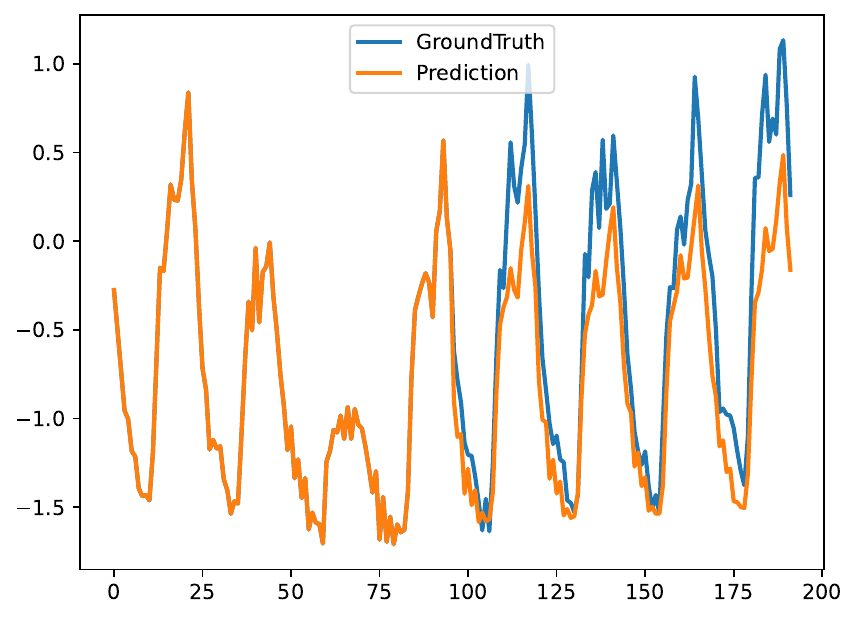}
    \caption{Prediction Visualization of TimeBase}
    \label{fig:power}
  \end{subfigure}
  \caption{Prediction visualization on the Electricity dataset (lookback=96, forecast horizon=96). We compare TimeCatcher with TimeMixer (state-of-the-art) and TimeBase (latest lightweight model). TimeCatcher more accurately captures abrupt load changes and sustained volatility patterns in the long-term horizon, demonstrating superior temporal fidelity under non-stationarity. }
  \label{fig:triple_horizontal}
\end{figure*}

\textbf{Evaluation Method.} Building upon previous works, we similarly employ the Mean Squared Error (MSE) and Mean Absolute Error (MAE) metrics to assess performance.

\textbf{Unified experimental setting.} It is worth noting that the results reported by previous baseline models cannot be directly compared due to inconsistencies in input sequence length and hyperparameter tuning strategies. To ensure fairness and rigor in the evaluation, we standardize the input lengths of all baseline models, carefully reimplement them under the same experimental setting, and report the average performance of three independent runs to mitigate randomness and provide more reliable comparisons.

\textbf{Implementation Details.} All the experiments are implemented in PyTorch (Paszke et al. 2019), and conducted on a single NVIDIA GeForce RTX 4090. We use the AdamW optimizer \cite{loshchilov2017fixing} with a learning rate selected based on the dataset. We train the model with L1 loss and adjust the hidden state dimensionality based on the length of the time series and the number of features to balance performance and efficiency.

\subsection{Main Result}
As shown in \textbf{Table 1}, TimeCatcher consistently achieves the lowest MSE and MAE across nearly all tasks, outperforming recent approaches such as TimeMixer (2024b) and TimeBase (2025). The improvements are particularly evident on datasets with complex temporal dynamics: for instance, the average MSE on the exchange rate dataset is reduced by \textbf{17.1\%} compared with the strongest baseline, while the average MAE on the solar energy dataset shows a \textbf{20.7\%} reduction. Additionally, the performance remains stable as the prediction length increases, indicating that the model maintains forecasting accuracy even at extended horizons where many existing approaches experience a rapid degradation in predictive quality.

A closer examination of results across individual datasets reveals consistent advantages in a variety of scenarios, including electricity load forecasting (ETT and ECL), meteorological prediction (Weather), and traffic flow modeling. On datasets characterized by high-frequency fluctuations and strong non-stationarity, such as exchange rates and solar energy, TimeCatcher achieves the best results in both short-term and long-term settings. Moreover, the method maintains competitive performance on more regular datasets, suggesting that its modeling capacity generalizes across different temporal structures and levels of complexity. Compared to recent multilayer perceptron-based predictors (such as DLinear and TimeBase), TimeCatcher demonstrates significant performance advantages across all benchmarks. In particular, it achieves performance improvements of over \textbf{50\%} on non-stationary series such as Solar Energy. This is due to our model's excellent ability to capture hidden information and model mutations.

\begin{table}[t]
\centering
\small
\renewcommand{\arraystretch}{1.2}
\setlength{\tabcolsep}{4.5pt}
\caption{Ablation Study on Trend Modeling, Latent Dynamic Encoding, and Volatility Enhancement components.  A check mark $\checkmark$ and an error mark \ding{55} indicate the presence or absence of certain components, respectively. \ding{172} denotes the official design of TimeCatcher.}
\begin{tabular}{c|c|c|c}
\toprule
Case & Trend Modeling & Latent Encoding & Volatility Enhancement\\
\midrule
\ding{172} & $\checkmark$ & $\checkmark$ & $\checkmark$  \\
\ding{173} & $\checkmark$ & $\checkmark$ & \ding{55} \\
\ding{174} & $\checkmark$ & \ding{55} & $\checkmark$ \\
\ding{175} & $\checkmark$ & \ding{55} & \ding{55} \\
\ding{176} & \ding{55} & $\checkmark$ & $\checkmark$ \\
\ding{177} & \ding{55} & $\checkmark$ & \ding{55} \\
\bottomrule
\end{tabular}
\label{tab:ablation}
\end{table}

The predictive behaviors shown in \textbf{Figure 3} provide additional evidence consistent with the quantitative findings. TimeCatcher’s prediction curves align closely with ground-truth observations, accurately reflecting both the amplitude and phase of the original series across multiple time intervals. In contrast, TimeMixer shows notable deviations during rapid transitions, and TimeBase exhibits visible lag and amplitude underestimation. These visual patterns correspond to the error metrics in \textbf{Table 1} and illustrate the model’s ability to capture dynamic variations and retain temporal fidelity across diverse forecasting conditions. Overall, TimeCatcher delivers accurate, robust, and efficient forecasting across diverse scenarios, effectively capturing complex dynamics and adapting to sudden changes.

\begin{figure*}[htbp]
  \centering
  \begin{subfigure}[b]{0.48\textwidth}
    \includegraphics[width=\linewidth]{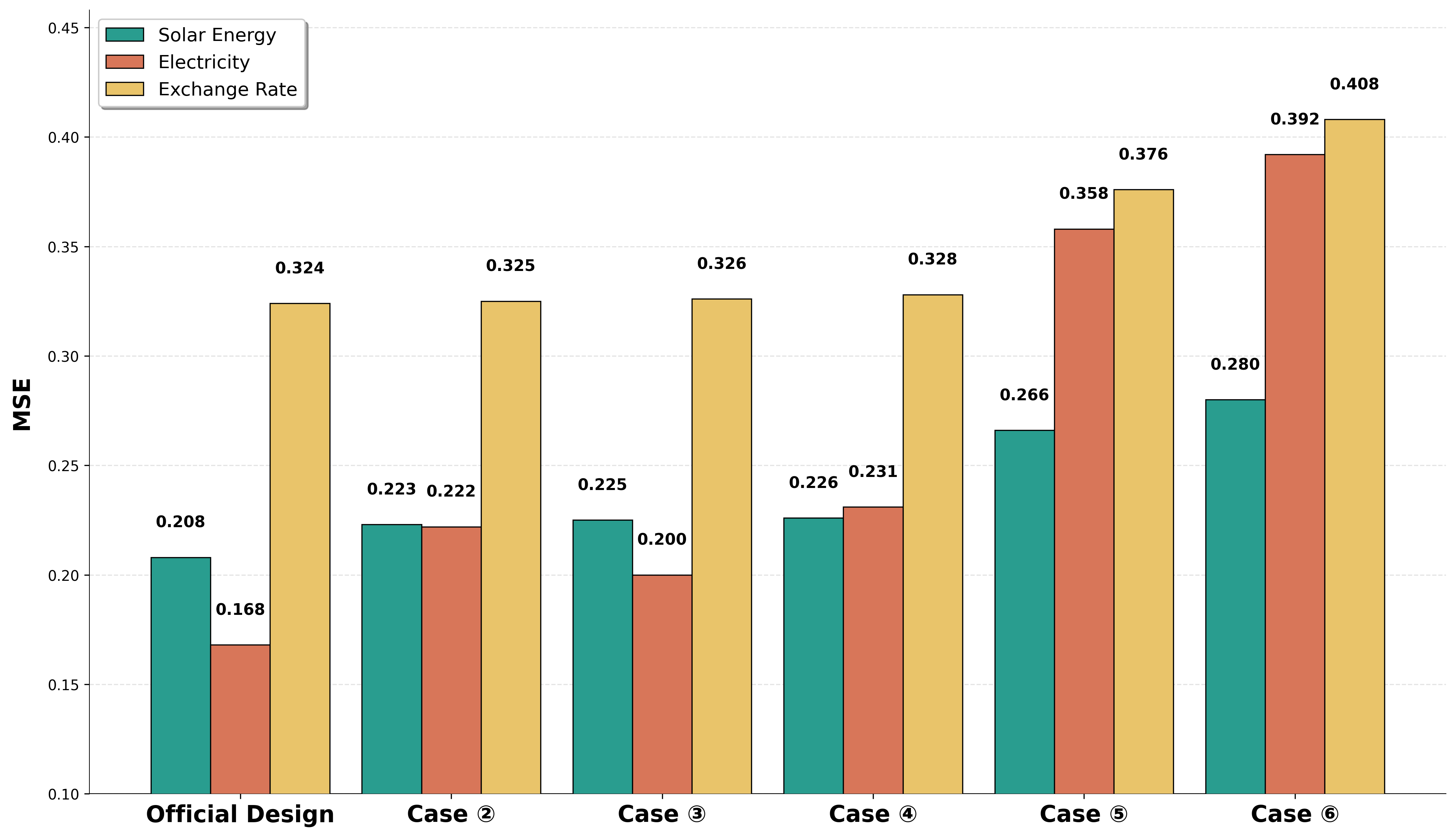}
    \caption{MSE of Different Cases}
    \label{fig:sub1}
  \end{subfigure}
  \hfill 
  \begin{subfigure}[b]{0.48\textwidth}
    \includegraphics[width=\linewidth]{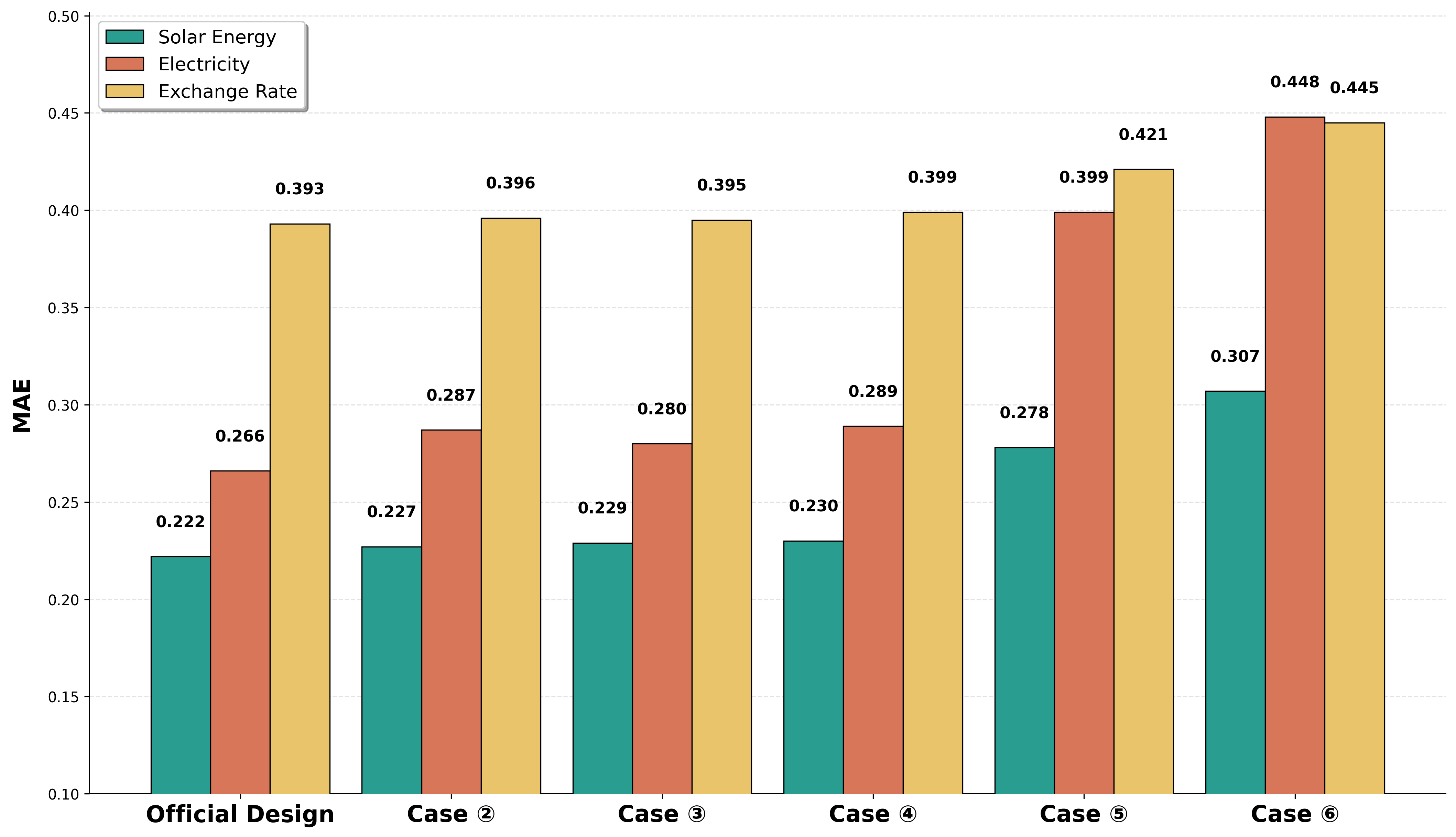}
    \caption{MAE of Different Cases}
    \label{fig:sub2}
  \end{subfigure}
  \caption{Ablation experiments show the predicted MSE and MAE values for different cases on the Solar Energy, Electricity, and Exchange Rate datasets. The MSE and MAE values for each case are the averages of the MSE and MAE values for four prediction lengths: $\{96, 192, 336, 720\}$. }
  \label{fig:main}
\end{figure*}

\begin{figure*}[htbp]
  \centering
  \begin{subfigure}[b]{0.32\textwidth}
    \includegraphics[width=\linewidth]{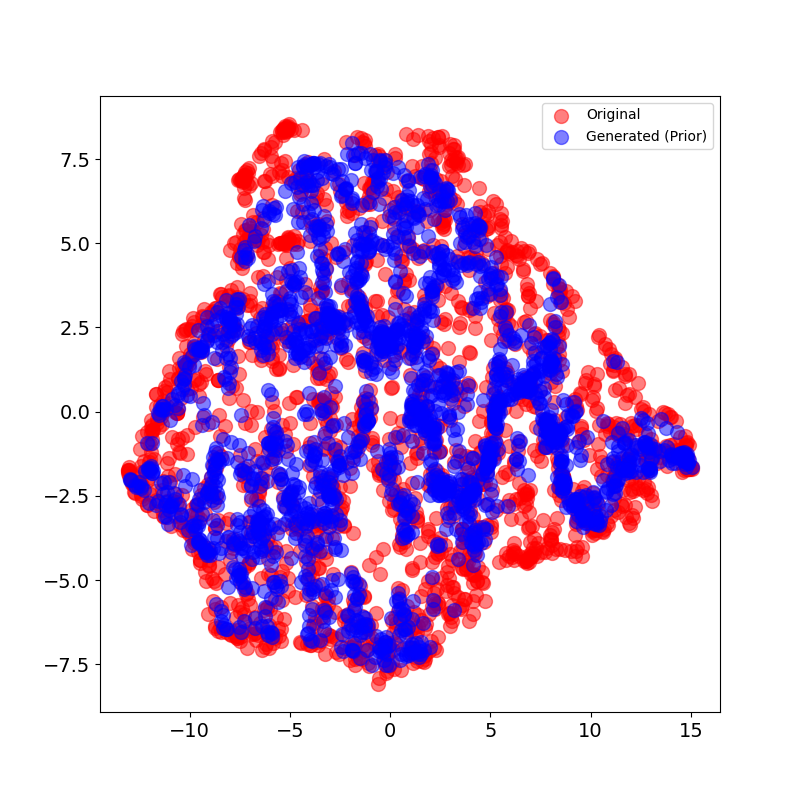}
    \caption{T-SNE plot of the ETTm1 dataset.}
    \label{fig:temp}
  \end{subfigure}
  \hfill
  \begin{subfigure}[b]{0.32\textwidth}
    \includegraphics[width=\linewidth]{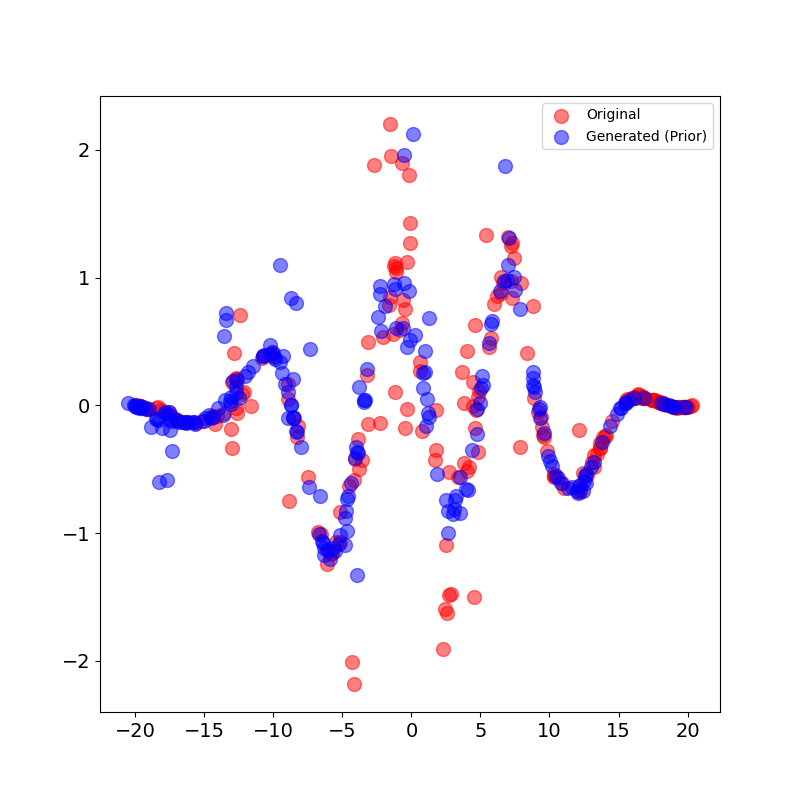}
    \caption{T-SNE plot of the Exchange Rate dataset.}
    \label{fig:mem}
  \end{subfigure}
  \hfill
  \begin{subfigure}[b]{0.32\textwidth}
    \includegraphics[width=\linewidth]{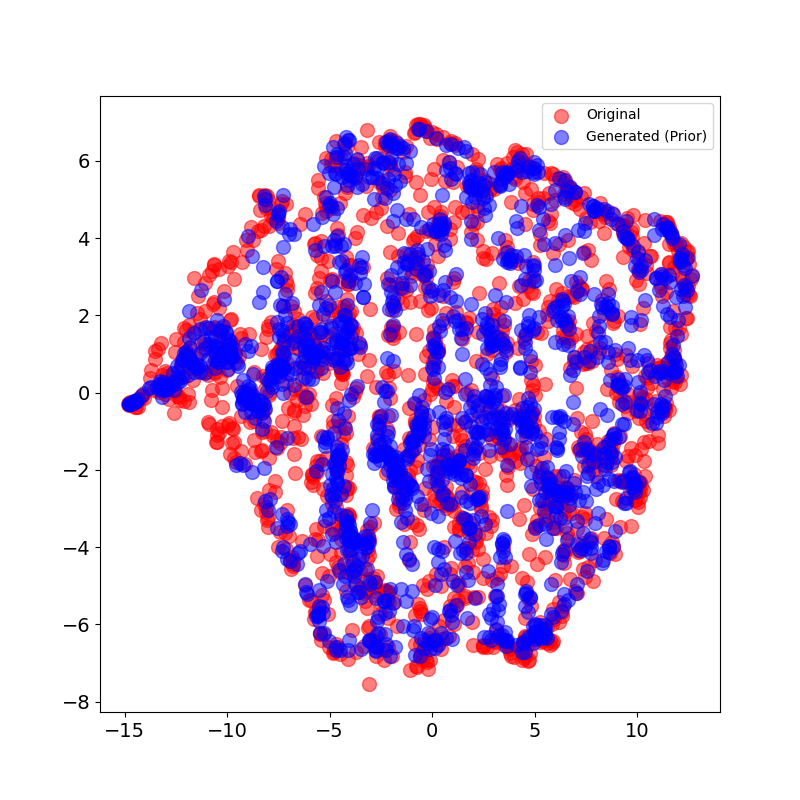}
    \caption{T-SNE plot of the Weather dataset.}
    \label{fig:power}
  \end{subfigure}
  \caption{T-SNE visualizations of latent representations in TimeCatcher on ETTm1 (left), Exchange Rate (center), and Weather (right). Red: original sequences; Blue: synthetic sequences from the variational encoder. The strong overlap and compact clustering indicate that TimeCatcher’s latent space preserves the underlying data distribution and captures structured temporal patterns. Additional visualizations are provided in Appendix B.}
  \label{fig:triple_horizontal}
\end{figure*}

\subsection{Ablation study}
To assess the contribution of each major component in TimeCatcher, we conduct ablation experiments by selectively removing the Trend Modeling, Latent Dynamic Encoding, and Volatility Enhancement modules, as summarized in \textbf{Table 2}. Six model variants are evaluated on three representative datasets — Solar Energy, Exchange Rate, and Electricity — with average MSE and MAE calculated across four prediction lengths. The complete design (Case \ding{172}) incorporates all components, while other cases examine the impact of omitting individual or multiple modules. At the same time, we also visualized the prediction results of all cases, which can be found in \textbf{Appendix A}.

\begin{figure*}[htbp]
  \centering
  \begin{subfigure}[b]{0.48\textwidth}
    \includegraphics[width=\linewidth]{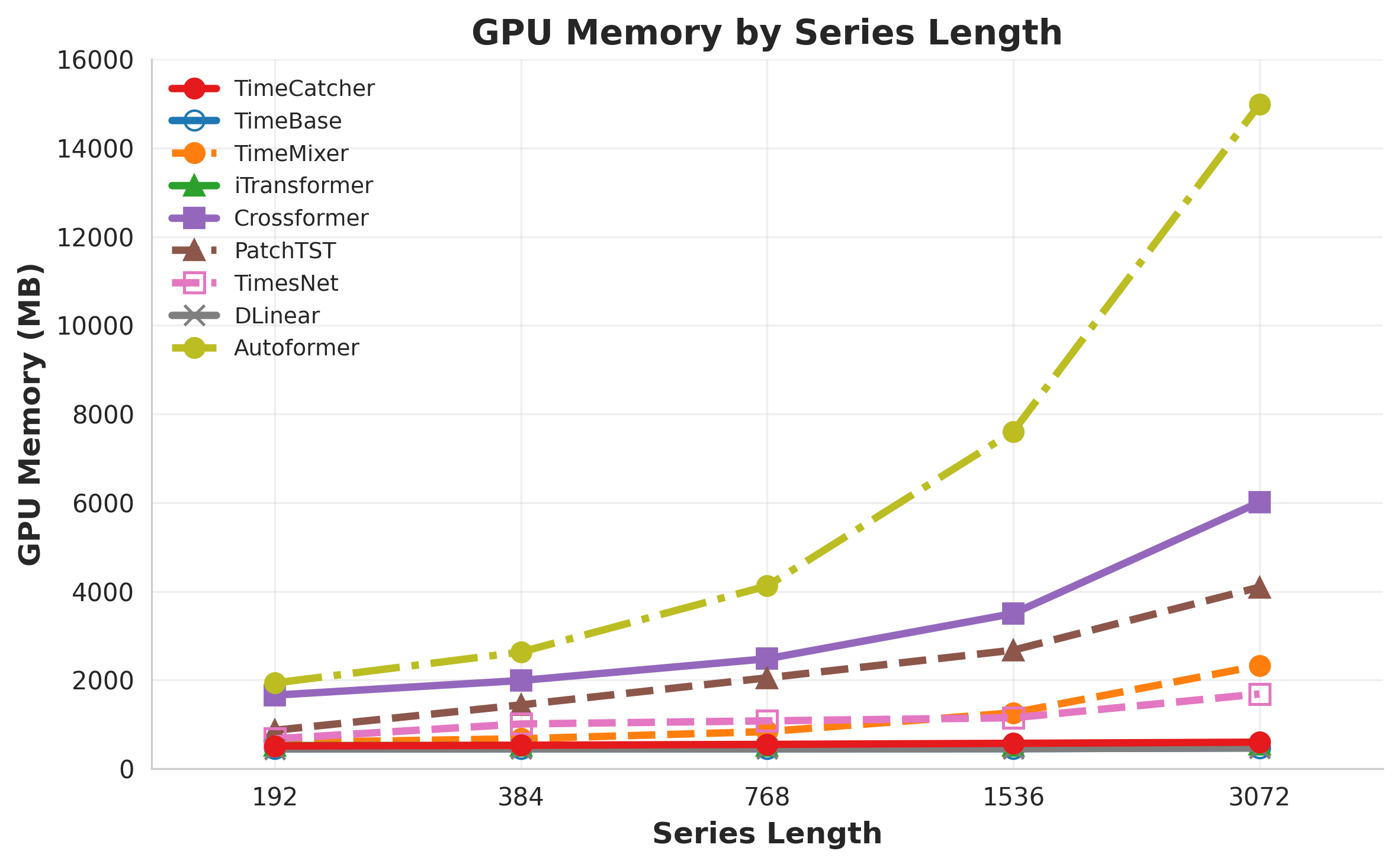}
    \caption{Memory Efficiency Analysis}
    \label{fig:sub1}
  \end{subfigure}-
  \hfill 
  \begin{subfigure}[b]{0.48\textwidth}
    \includegraphics[width=\linewidth]{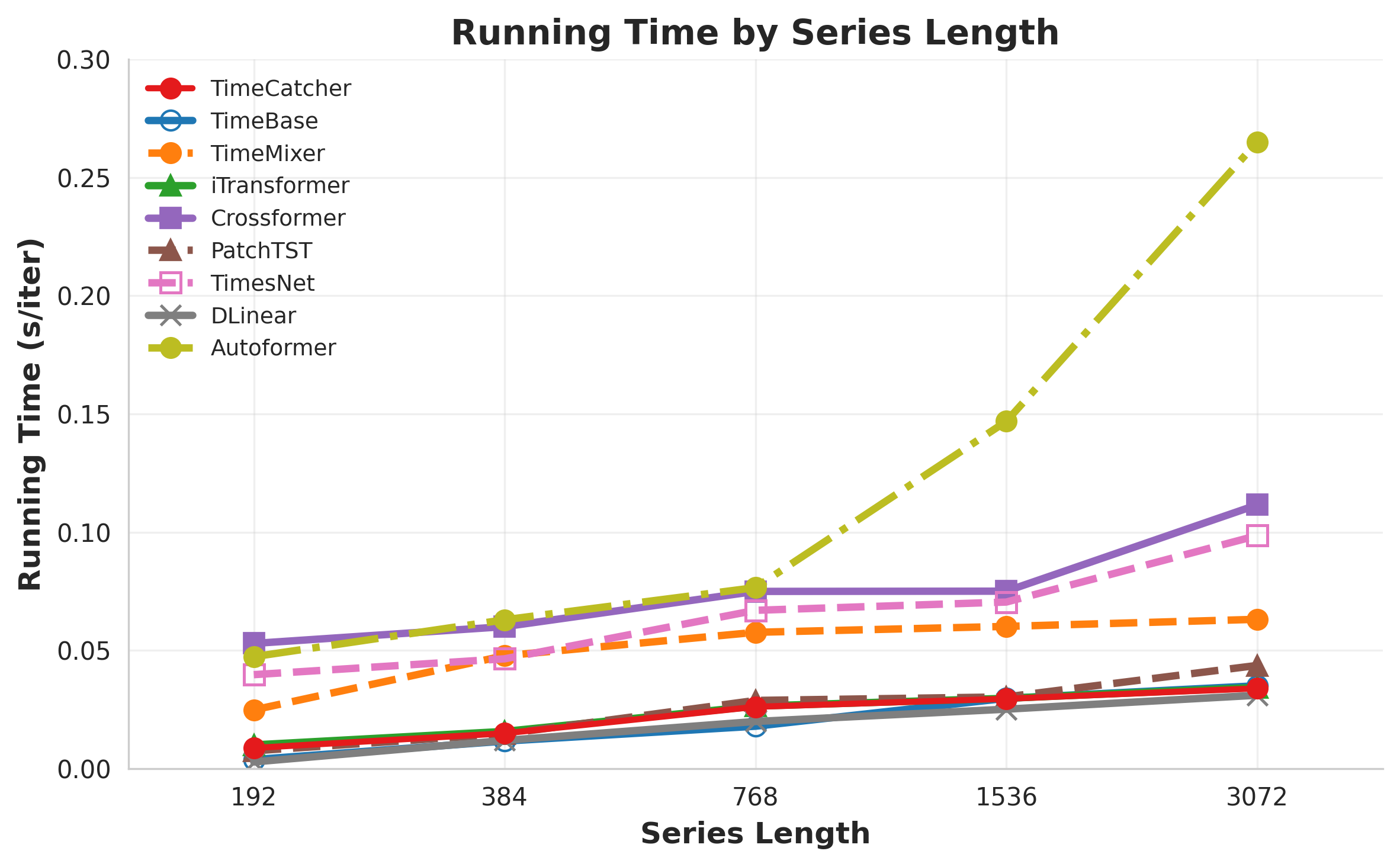}
    \caption{Running Time Efficiency Analysis}
    \label{fig:sub2}
  \end{subfigure}
  \caption{Efficiency analysis in both GPU memory and running time. The results are recorded on the ETTh1 dataset with batch size as 16. The running time is averaged from $10^2$ iterations.}
  \label{fig:main}
\end{figure*}

The results in \textbf{Figure 4} show that removing any single component leads to performance degradation across all datasets. Excluding the Volatility Enhancement module (Case \ding{173}) causes a noticeable increase in both MSE and MAE, particularly on the Exchange Rate dataset, suggesting that the ability to capture abrupt temporal changes is closely linked to predictive accuracy. Similarly, removing the Latent Dynamic Encoding component (Case \ding{174}) results in elevated errors, especially on the Solar Energy dataset, where latent temporal dependencies play a significant role. The absence of the Trend Modeling branch (Case \ding{176} and \ding{177}) produces the largest degradation overall, with MSE values exceeding 0.40 on the Exchange Rate dataset and MAE rising above 0.44, indicating that direct modeling of raw series information is fundamental to the forecasting task.

When multiple modules are removed simultaneously (Case \ding{175}–\ding{177}), the degradation becomes more pronounced, and the performance gap relative to the full model widens. This cumulative decline highlights the complementary roles of the three components: source prediction ensures a robust baseline representation, latent information extraction enriches temporal context, and volatility modeling enhances sensitivity to rapid changes. Together, these findings confirm that the full architecture achieves its predictive performance through the joint contribution of all three modules, with each component addressing distinct aspects of temporal dynamics.

\subsection{Model Analysis}
The latent representation module in TimeCatcher is implemented based on a variational autoencoder (VAE) framework, where the stochastic nature of latent variables plays a crucial role during training. In each forward pass, the evidence lower bound (ELBO) introduces a sampling term that treats the latent variable $\mathrm{z}$, as a random vector, and model parameters are optimized with respect to samples from $q_{\phi}(\mathrm{z} \mid \mathrm{x} )$. This process refines the conditional distribution $p_{\theta}(\mathrm{x} \mid \mathrm{z})$, allowing the dynamic statistical properties present in the training data to be implicitly parameterized and inherited by the model. During inference, the latent variable sequence is further decomposed into trend and seasonal components, which are linearly projected into future steps to produce the prediction path $\hat{\mathrm{z}}$. Since the weights of these linear layers have been shaped under ELBO regularization, the predictions naturally encode dynamic priors learned during training. As shown in the ablation results (\textbf{Figure 4}), removing this stochastic pathway results in an MSE increase of 0.018 on the Exchange dataset and 0.054 on the Solar dataset, indicating that the latent space contributes directly to forecasting accuracy through the retention of dynamic statistical information.

The distributional properties of the latent space are further examined using t-SNE visualizations (\textbf{Figure 5}). Across all datasets, reconstructed samples generated from the latent space exhibit strong overlap with the original data points, indicating that the learned latent manifold faithfully preserves the underlying structure and geometry of the input sequences. No clear cluster separation is observed between real and synthetic samples, suggesting that the VAE avoids mode collapse and effectively captures the diversity of the original data distribution. Moreover, the latent space distribution is more compact than the raw time series, reflecting the removal of noise and irrelevant variability while retaining essential temporal patterns. These results collectively show that TimeCatcher transforms dynamic characteristics into a structured latent representation that supports stable and generalizable long-term forecasting. Overall, these analyses show that the latent module effectively links past dynamics with future predictions, allowing TimeCatcher to capture and utilize temporal structures.

\subsection{Efficiency Analysis}
As quantitatively demonstrated in \textbf{Figure 6}, we conduct a comprehensive efficiency analysis by comparing the GPU memory consumption and running time during training against latest state-of-the-art models. The results reveal that TimeCatcher maintains significantly superior efficiency across varying series lengths (192 to 3072). Specifically, in the memory efficiency analysis (\textbf{Figure 5a}), TimeCatcher's memory usage growth rate is much lower than that of other models, and the total memory usage always remains below 600MB, which demonstrates better scalability. Meanwhile, in running time analysis (\textbf{Figure 5b}), TimeCatcher consistently achieves the lowest time consumption per iteration among all compared methods. This efficiency advantage, combined with its consistent state-of-the-art performance on long-term prediction tasks, makes TimeCatcher particularly suitable for resource-constrained environments and real-time deployment scenarios.

\section{Conclusion}
In this work, we proposed TimeCatcher, a novel volatility-aware variational forecasting framework that integrates deterministic trend modeling, latent representation learning, and fluctuation-sensitive enhancement into a unified MLP-based architecture. By combining a VAE encoder with a lightweight volatility emphasis module, TimeCatcher effectively captures both smooth long-term dynamics and abrupt non-stationary variations, achieving consistent state-of-the-art performance across diverse real-world benchmarks. Extensive experiments demonstrate its robustness in highly volatile domains such as financial markets, energy demand, and solar generation. Beyond these tested scenarios, the framework also holds strong potential for broader applications where sudden fluctuations are critical, including web traffic monitoring and anomaly detection. Overall, TimeCatcher highlights the importance of hybrid designs that balance efficiency, interpretability, and adaptability. Future work may extend this framework to online learning in multimodal signal environments and to a wider range of applications requiring prediction under uncertainty.


\clearpage

\bibliographystyle{ACM-Reference-Format}
\bibliography{TimeCatcher_ref}
\newpage
\appendix
\section{Full Results}
Due to space limitations, the main text does not include detailed prediction curve comparisons. This appendix provides a visual comparison of the predicted values and true values for different ablation cases on the ECL dataset under the input-96-prediction-96 setting, as shown in \textbf{Figure 7}. This provides a qualitative and intuitive demonstration of the role of each core component.

\textbf{Figure 7} clearly shows that the prediction curve (orange) of the complete model (Case \ding{172}, official design) fits the true value curve (blue) most closely, accurately tracking the fluctuation trend and local details of the series. In sharp contrast, the prediction results of each ablation case show varying degrees of performance degradation. Specifically, the model with the latent information capture module removed (such as Case \ding{175}) performs poorly in predicting fine fluctuations in the series, resulting in an overly smooth prediction curve. The model with the fluctuation information capture module removed (such as Case \ding{174}) struggles to adapt to sudden changes in the data, exhibiting significant prediction bias near the mutation point. When the source data prediction module is disabled (such as Case \ding{177}), the prediction results even deviate from the overall trend of the true series.

This visual analysis provides intuitive support for the quantitative ablation study in the main text. The differences in the prediction curves' morphology—such as phase deviation, amplitude distortion, and trend misidentification—are highly consistent with the theoretical capabilities of each component in modeling temporal dependencies, hidden dynamics, and volatility. These qualitative results demonstrate that the overall design of TimeCatcher is essential, and that its components work together in a complementary manner to ensure the model's robust prediction capabilities on complex real-world data.

\section{T-SNE Plots of Datasets}
Due to the length limitation of the main text, we summarize the t-SNE distribution of the remaining datasets in \textbf{Figure 6}. This figure presents a comparative visualization of the two-dimensional t-SNE projections for five distinct datasets: ETTh1, ETTh2, ETTm2, Electricity, and Traffic. In each subplot, the original data points are marked in red, while their generated counterparts are depicted in blue, allowing for an intuitive side-by-side comparison of their underlying manifold structures. The uniform layout and consistent color scheme across all subplots facilitate a clear and direct assessment of distributional similarities and differences.

The visualization reveals that, across all five datasets, the distributions of the generated data (blue) exhibit a remarkable degree of overlap with the original data (red). This high level of structural congruence suggests that our model effectively captures and preserves the essential topological characteristics of the original time series in the latent space. Such alignment is a strong indicator of the generation quality and provides visual evidence supporting the model's ability to produce realistic and structurally faithful synthetic data, as discussed in the main text.

\clearpage

\begin{figure*}[htbp]
  \centering
  \begin{subfigure}[b]{0.32\textwidth}
    \includegraphics[width=\linewidth]{ab0.pdf}
    \caption{Official Design}
    \label{fig:sub1}
  \end{subfigure}
  \hfill
  \begin{subfigure}[b]{0.32\textwidth}
    \includegraphics[width=\linewidth]{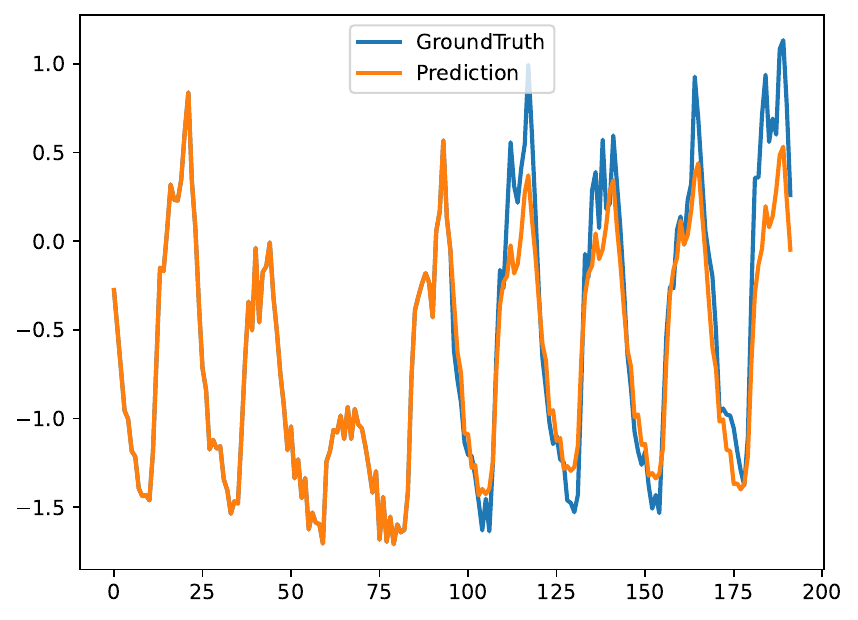}
    \caption{Case \ding{173}}
    \label{fig:sub2}
  \end{subfigure}
  \hfill
  \begin{subfigure}[b]{0.32\textwidth}
    \includegraphics[width=\linewidth]{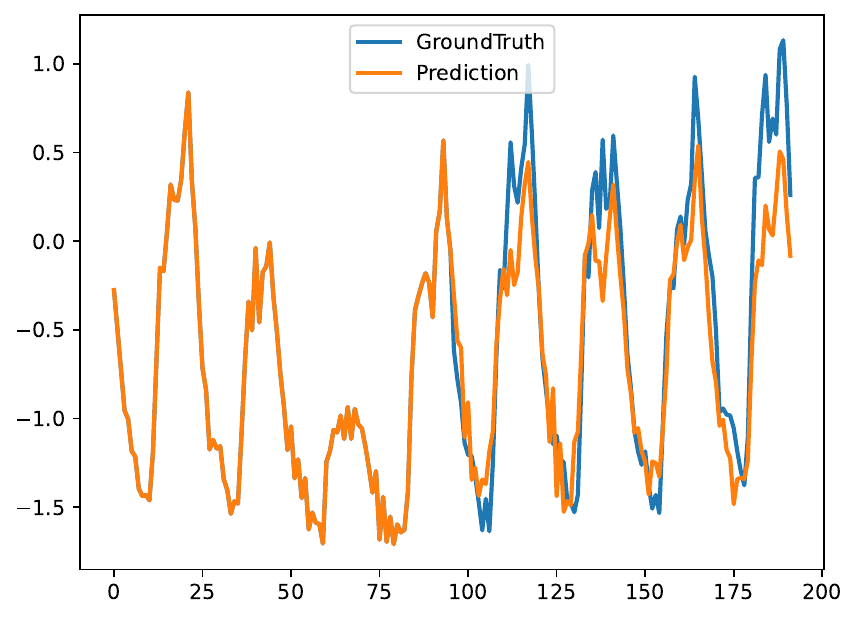}
    \caption{Case \ding{174}}
    \label{fig:sub3}
  \end{subfigure}
  
  \vspace{0.4cm}
  \begin{subfigure}[b]{0.32\textwidth}
    \includegraphics[width=\linewidth]{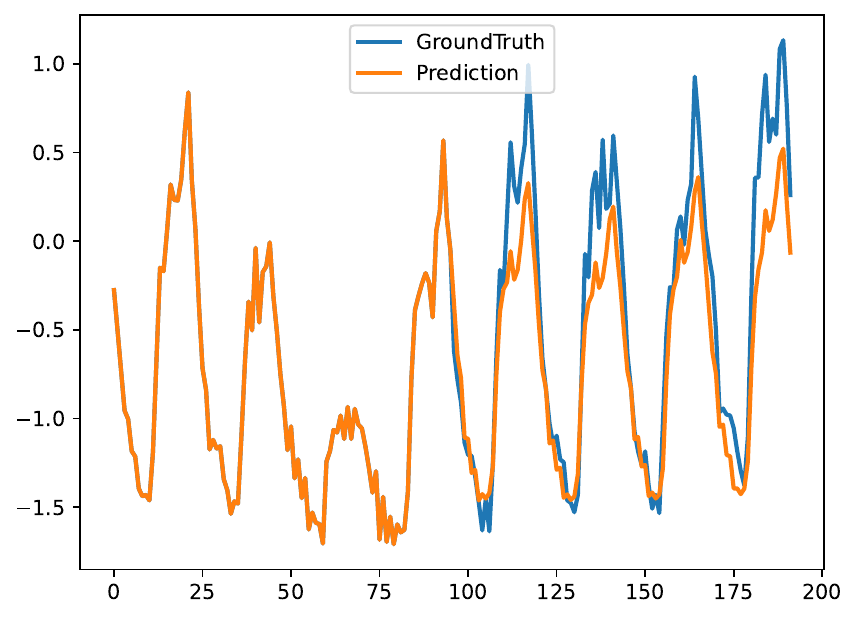}
    \caption{Case \ding{175}}
    \label{fig:sub4}
  \end{subfigure}
  \hfill
  \begin{subfigure}[b]{0.32\textwidth}
    \includegraphics[width=\linewidth]{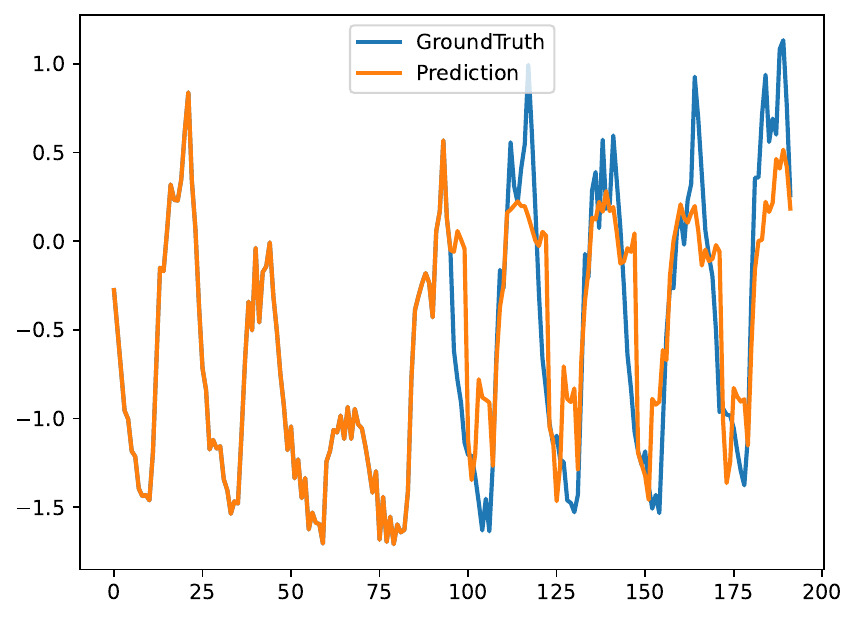}
    \caption{Case \ding{176}}
    \label{fig:sub5}
  \end{subfigure}
  \hfill
  \begin{subfigure}[b]{0.32\textwidth}
    \includegraphics[width=\linewidth]{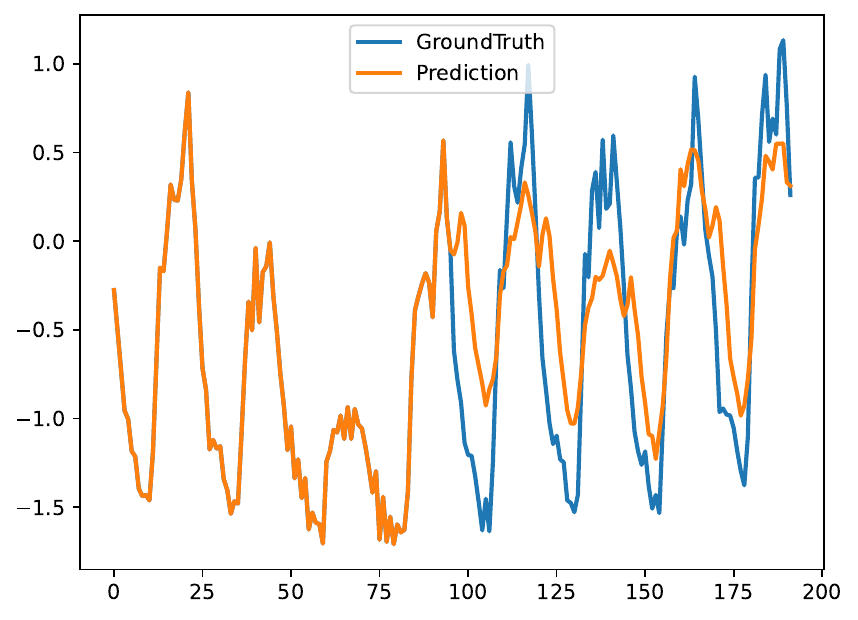}
    \caption{Case \ding{177}}
    \label{fig:sub6}
  \end{subfigure}
  
  \caption{Comparative visualization of predictions for different combinations under the input-96-prediction-96 setting of the ECL dataset.}
  \label{fig:2x3_grid}
\end{figure*}

\begin{figure*}[t]
    \begin{subfigure}[b]{0.3\textwidth}
        \includegraphics[width=\textwidth]{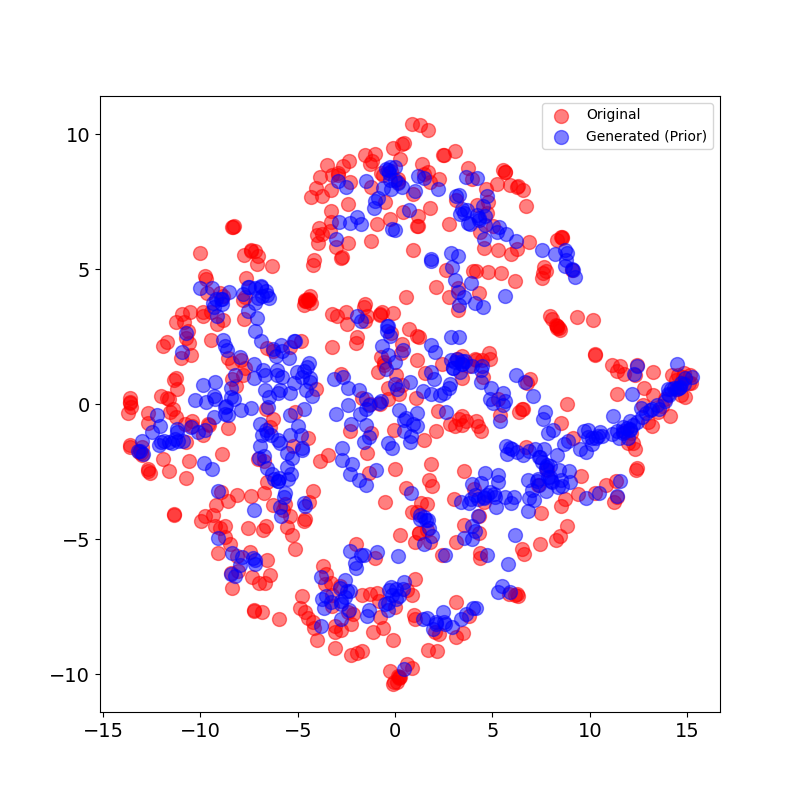}
        \caption{T-SNE plot of the ETTh1 dataset.}
    \end{subfigure}
    \begin{subfigure}[b]{0.3\textwidth}
            \includegraphics[width=\textwidth]{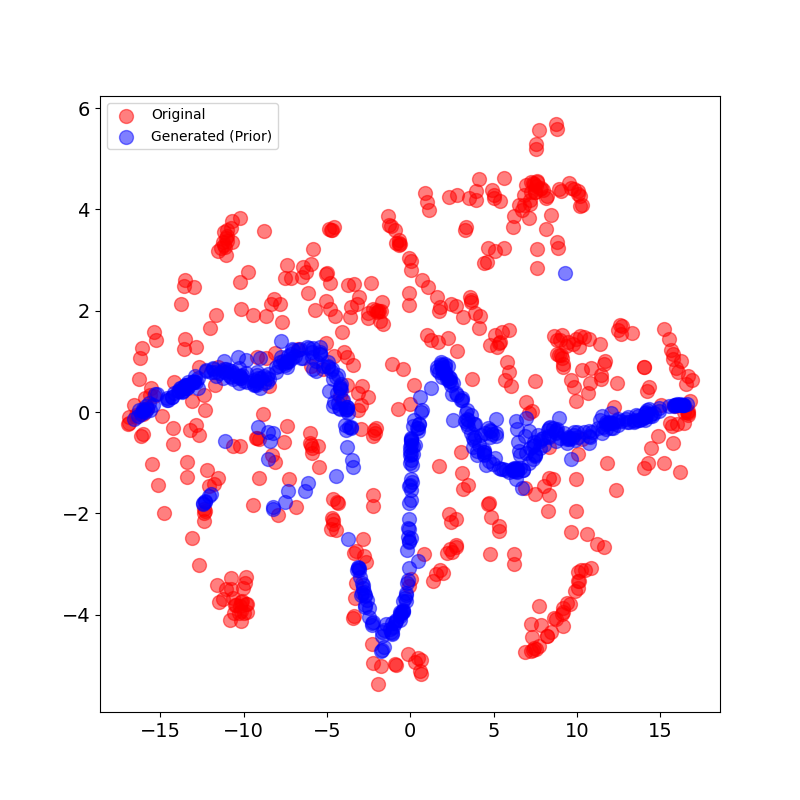}
        \caption{T-SNE plot of the ETTh2 dataset.}
    \end{subfigure}
    
    \vspace{10pt}
    \begin{subfigure}[b]{0.3\textwidth}
            \includegraphics[width=\textwidth]{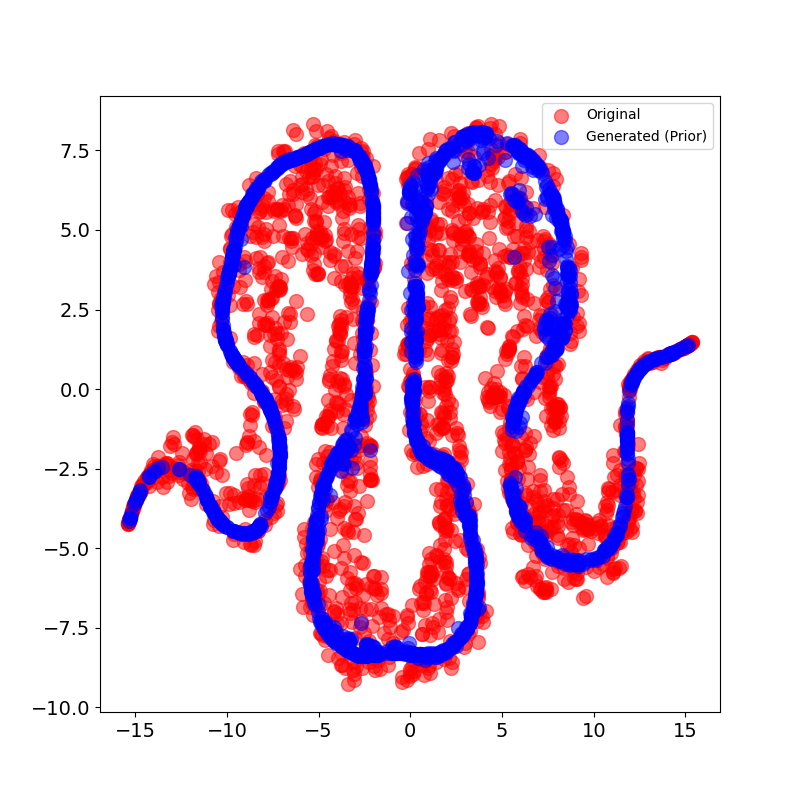}
        \caption{T-SNE plot of the ETTm2 dataset.}
    \end{subfigure}
    \hfill
    \begin{subfigure}[b]{0.3\textwidth}
        \includegraphics[width=\textwidth]{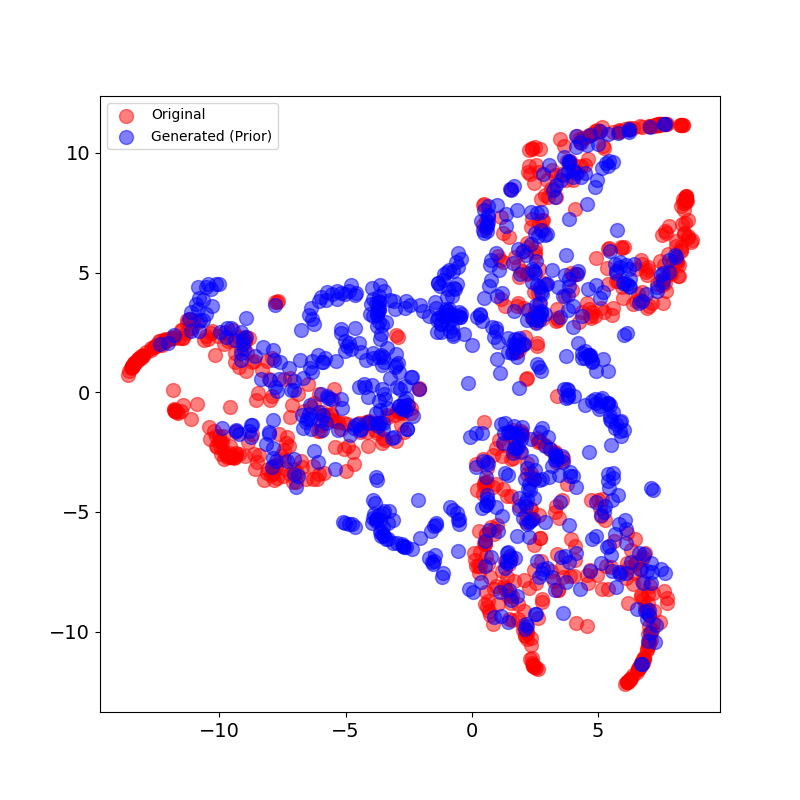}
        \caption{T-SNE plot of the Electricity dataset.}
    \end{subfigure}
    \hfill
    \begin{subfigure}[b]{0.3\textwidth}
        \includegraphics[width=\textwidth]{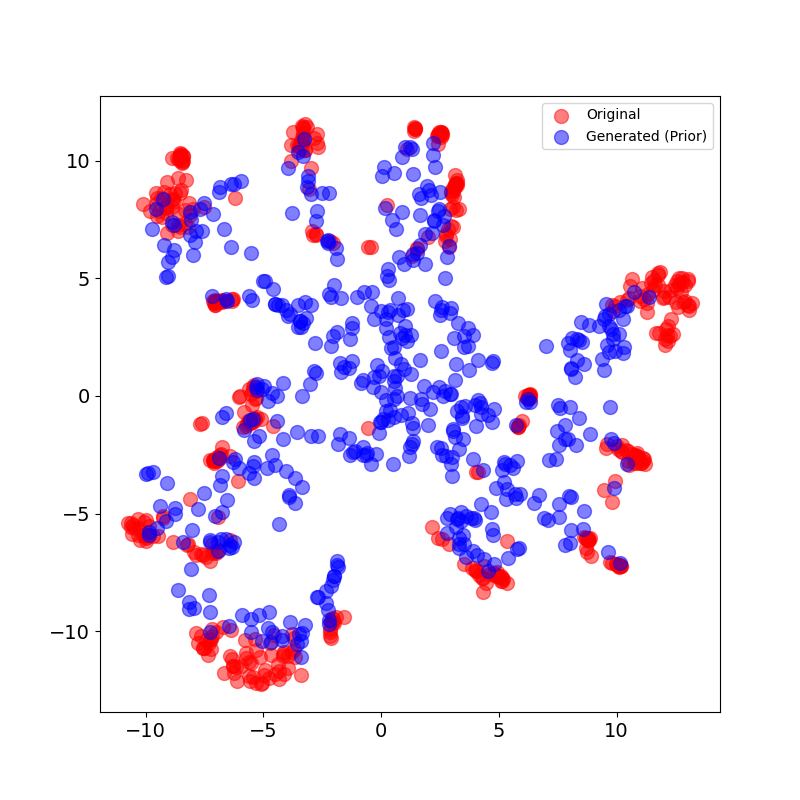}
        \caption{T-SNE plot of the Traffic dataset.}
    \end{subfigure}
    
    \caption{T-SNE plots of the remaining datasets (ETTh1, ETTh2, ETTm2, Electricity, Traffic).}
    \label{fig:5figs}
\end{figure*}
\end{document}